%% file: main.tex
\documentclass[twoside]{article}

\usepackage[pagebackref=true,breaklinks=true,colorlinks,]{hyperref}

\usepackage[toc,page,header]{appendix}

\usepackage{minitoc}

\usepackage[accepted]{aistats2023/aistats2023}
\usepackage[ruled,vlined, noend]{algorithm2e}
\input{contents/shared_macros}

\theoremstyle{plain}
\newtheorem{theorem}{Theorem}
\newtheorem{proposition}{Proposition}

\newtheorem{lemma}{Lemma}

\theoremstyle{definition}
\newtheorem{definition}{Definition}
\newtheorem{assumption}{Assumption}

\theoremstyle{remark}

\usepackage{url}
\begin{document}

%

%

\twocolumn[

\aistatstitle{Group Distributionally Robust Reinforcement Learning with \\Hierarchical Latent Variables}

\centering{
\textbf{
Mengdi Xu$^{1}$, Peide Huang$^{1}$, Yaru Niu$^{1}$, Visak Kumar$^{2}$, Jielin Qiu$^{1}$
}
}

\centering{
\textbf{
Chao Fang$^{2}$, Kuan-Hui Lee$^{2}$, Xuewei Qi,
Henry Lam$^{3}$ , Bo Li$^{4}$, Ding Zhao$^{1}$
}
}

\aistatsaddress{ } 
]

\begin{abstract}
One key challenge for multi-task Reinforcement learning (RL) in practice is the absence of task indicators. Robust RL has been applied to deal with task ambiguity, but may result in over-conservative policies. To balance the worst-case (robustness) and average performance, we propose Group Distributionally Robust Markov Decision Process (GDR-MDP), a flexible hierarchical MDP formulation that encodes task groups via a latent mixture model. GDR-MDP identifies the optimal policy that maximizes the expected return under the worst-possible qualified belief over task groups within an ambiguity set. We rigorously show that GDR-MDP's hierarchical structure improves distributional robustness by adding regularization to the worst possible outcomes. We then develop deep RL algorithms for GDR-MDP for both value-based and policy-based RL methods. Extensive experiments on Box2D control tasks, MuJoCo benchmarks, and Google football platforms show that our algorithms outperform classic robust training algorithms across diverse environments in terms of robustness under belief uncertainties. Demos are available on our project page (\url{https://sites.google.com/view/gdr-rl/home}).
\end{abstract}


\doparttoc 
\faketableofcontents 

\input{contents/content_main}

\bibliography{citation}
\bibliographystyle{unsrt}

\newpage

\onecolumn

\appendix
\input{contents/appendix}

\end{document}

%% file: contents/shared_macros.tex
\usepackage{wrapfig}
\usepackage{subcaption}

\usepackage{amsmath}
\usepackage{amssymb}
\usepackage{mathtools}
\usepackage{amsthm}
\usepackage{booktabs,multirow,multicol}

\usepackage{amsfonts}
\usepackage{bbm}
\usepackage{mathrsfs}
\usepackage{times}
\usepackage{helvet}
\usepackage{courier}
\usepackage{graphicx}
\usepackage{color}
\usepackage{colortbl}

\newcommand{\kibitz}[2]{\ifnum\Comments=0\textcolor{#1}{#2}\fi}
\definecolor{darkgreen}{rgb}{0,0.5,0}

\usepackage{amsmath,amsfonts,bm}

\def\eqref#1{equation~\ref{#1}}

\def\1{\bm{1}}

\def\rvp{{\mathbf{p}}}

\def\ermP{{\textnormal{P}}}

\def\ermR{{\textnormal{R}}}

\def\mA{{\bm{A}}}

\DeclareMathAlphabet{\mathsfit}{\encodingdefault}{\sfdefault}{m}{sl}
\SetMathAlphabet{\mathsfit}{bold}{\encodingdefault}{\sfdefault}{bx}{n}

\def\gA{{\mathcal{A}}}

\def\gC{{\mathcal{C}}}
\def\gD{{\mathcal{D}}}

\def\gL{{\mathcal{L}}}
\def\gM{{\mathcal{M}}}

\def\gS{{\mathcal{S}}}

\def\gX{{\mathcal{X}}}

\def\gZ{{\mathcal{Z}}}

\DeclareMathOperator*{\argmin}{arg\,min}

\DeclareMathOperator{\sign}{sign}


%% file: contents/content_main.tex
\vspace{-0.15in}
\section{Introduction}
\vspace{-0.1in}
Reinforcement learning (RL) has demonstrated extraordinary capabilities in sequential decision-making, even for handling multiple tasks \cite{mnih2013playing, kober2013reinforcement, kirk2021survey, finn2017model}.
With policies conditioned on accurate task-specific contexts, RL agents could perform better than ones without access to context information \cite{steimle2021multi, sodhani2021multi}.
However, one key challenge for contextual decision-making is that, in real deployments, RL agents may only have incomplete information about the task to solve.  
In principle, agents could adaptively infer the latent context with data collected across an episode, and prior knowledge about tasks \cite{wilson2007multi, rakelly2019efficient, hausman2018learning}.
However, the context estimates may be inaccurate \cite{xie2022robust,sharma2019robust} due to limited interactions, poorly constructed inference models, or intentionally injected adversarial perturbations. 
Blindly trusting the inferred context and performing context-dependent decision-making may lead to significant performance drops or catastrophic failures in safety-critical situations.
Therefore, in this work, we are motivated to study the problem of \textit{robust decision-making under the task estimate uncertainty}. 

\vspace{-0.05in}
Prior works about robust RL involve optimizing over the worst-case qualified elements within one uncertainty set \cite{nilim2005robust,iyengar2005robust}. 
Such robust criterion assuming the worst possible outcome may lead to \textit{overly conservative policies}, or even training instabilities \cite{zhang2020stability,yu2021robust, huang2022robust}. For instance, an autonomous agent trained with robust methods may always assume the human driver is aggressive regardless of recent interactions and wait until the road is clear, consequently blocking the traffic.
Therefore, balancing the robustness against task estimate uncertainties and the performance when conditioned on the task estimates is still an open problem. 
We provide one solution to address the above problem by \textit{modeling the commonly existing similarities between tasks under distributionally robust Markov Decision Process (MDP) formulations}.

\vspace{-0.05in}

\begin{figure}[t]
    \centering
    \includegraphics[width=0.98\linewidth]{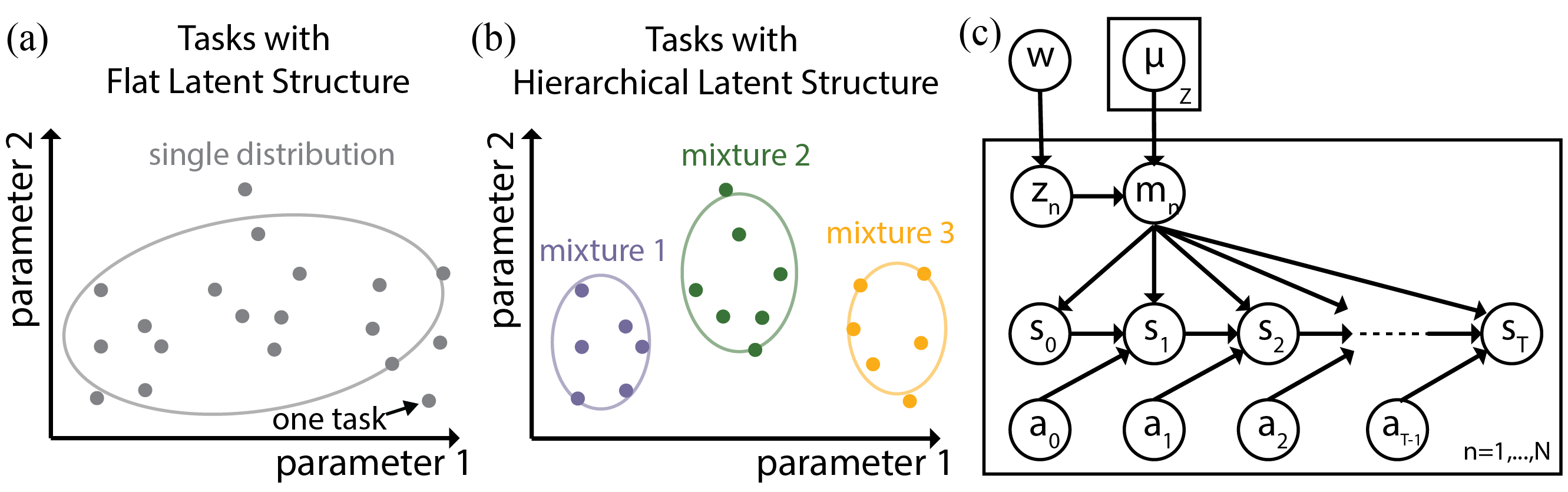}
    \vspace{-0.1in}
    \caption{Illustration examples when modeling tasks with a flat latent structure that uses one distribution for all tasks as in (a), and a hierarchical latent structure that clusters tasks to different mixtures as in (b). The graphical model with a hierarchical latent structure for both GDR-MDP and HLMDP is shown in (c). At episode $n$, a mixture $z_n$ is first sampled from a prior distribution $w$. An MDP $m$ is then sampled according to $\mu_{z_n}(m)$ and controls the dynamics of the $n$'th episode.
    }
    \label{fig:illustration}
    \vspace{-0.05in}
\end{figure}

Each task is typically represented by a unique combination of parameters or a multi-dimensional context in multi-task RL. We argue that some parameters are more important than others in terms of affecting the environment dynamics model and thus tasks can be properly clustered into mixtures according to the more crucial parameters as in Figure \ref{fig:illustration}~(a) and (b). 
However, existing robust MDP formulations \cite{nilim2005robust} lack the capacity to model task groups, or equivalently, task subpopulations.
Thus the effect of task subpopulations on the policy's robustness is unexplored.
In this paper, we show that the task subpopulations help balance the worst-case performance (robustness) and average performance under conditions (Section~\ref{sec:analysis}). 

\vspace{-0.05in}

In contrast to prior work \cite{xie2022robust} that leverages point estimates of latent contexts, we take a probabilistic point of view and represent the task subpopulation estimate with a belief distribution. 
Holding a belief of the task subpopulation, which is the high-level latent variable, helps leverage the prior distributional information of task similarities. It also naturally copes with distributionally robust optimization by optimizing w.r.t. the worst-possible belief distribution within an ambiguity set. 
We consider an adaptive setting in line with system identification methods \cite{yu2017preparing}, where the belief is initialized as a uniform distribution and then updated during one episode.
Our problem formation is related to the ambiguity modeling \cite{etner2012decision} inspired by human's bounded rationality to approximate and handle distributions, which has been studied in behavioral economics \cite{ellsberg1961risk, machina2014ambiguity} yet has not been widely acknowledged in RL.

We highlight our main contributions as follows:
\vspace{-0.1in}
\begin{enumerate}
    \vspace{-0.05in}
    \item We formulate Hierarchical-Latent MDP (HLMDP) (Section~\ref{sec:HLMDP}), which utilizes a mixture model over MDPs to encode task subpopulations. 
    HLMDP has a high-level latent variable $z$ as the mixture, and a low-level $m$ to represent tasks (Figure~\ref{fig:illustration} (c)). 
    \vspace{-0.05in}
    \item We introduce the Group Distributionally Robust MDP (GDR-MDP) in Section~\ref{sec:GDR-MDP} to handle the over-conservative problem, which formulates the robustness w.r.t. the ambiguity of the adaptive belief $b(z)$ over mixtures. GDR-MDP builds on distributionally robust optimization \cite{rahimian2019distributionally, kuhn2019wasserstein} and HLMDP to leverage rich distributional information.
    \vspace{-0.05in}
    \item We show the convergence property of GDR-MDP in the infinite-horizon case. We find that the hierarchical latent structure helps restrict the worst-possible outcome within the ambiguity set and thus helps generate less conservative policies with higher optimal values.
    \vspace{-0.05in}
    \item We design robust deep RL training algorithms based on GDR-MDP by injecting perturbations to beliefs stored in the data buffer. We empirically evaluate in three environments, including robotic control tasks and google research football tasks. Our results demonstrate that our proposed algorithms outperform baselines in terms of robustness to belief noise.
\end{enumerate}


\section{Related Work}
\label{sec:related_work}
\vspace{-0.1in}

\paragraph{Robust RL and Distributionally Robust RL.}
RL's vulnerability to uncertainties has attracted large efforts to design proper robust MDP formulations accounting for uncertainties in MDP components \cite{nilim2005robust,iyengar2005robust, wiesemann2013robust, osogami2015robust, tessler2019action, zhang2020robust}.
Existing robust deep RL algorithms \cite{moos2022robust,klibanoff2005smooth,foerster2017learning, pinto2017robust, Zhang_etal_robust_model_uncertainity, osogami2015robust} are shown to generate robust policies with promising results in practice.
However, it is also known that robust RL that optimizes over the worst-possible elements in the uncertainty set may generate over-conservative policies by trading average performance for robustness and may even lead to training instabilities \cite{huang2022robust}. 
In contrast, distributionally robust RL \cite{xu2010distributionally, yu2015distributionally, smirnova2019distributionally, grand2020first, zhou2021finite, nakao2021distributionally, sinha2020formulazero, Delage2010DistributionallyRO} assumes that the \textit{distribution} of uncertain components (such as transition models) is partially/indirectly observable.
It builds on distributionally robust optimization
\cite{rahimian2019distributionally, kuhn2019wasserstein} which optimizes over the worst possible distribution within the ambiguity set. 
Compared with common robust methods, distributionally robust RL embeds prior probabilistic information and generates less conservative policies with carefully calibrated ambiguity sets \cite{xu2010distributionally}. We aim to propose distributionally robust RL formulations and training algorithms to handle task estimate uncertainties while maintaining a trade-off between robustness and performance.

One relevant work is the recently proposed distributionally robust POMDP \cite{nakao2021distributionally} which maintains a belief over states and finds the worst possible transition model distribution within an ambiguity set. We instead hold a belief over task mixtures and find the worst possible belief distribution.
\cite{sinha2020formulazero} also maintains a belief distribution over tasks but models tasks with a flat latent structure. Moreover, \cite{sinha2020formulazero} achieves robustness by optimizing at test-time, while we aim to design robust training algorithms to save computation during deployment.

\vspace{-0.1in}
\paragraph{RL with Task Estimate Uncertainty.} 
Inferring the latent task as well as utilizing the estimates in decision-making have been explored under the framework of Bayesian-adaptive MDPs \cite{ghavamzadeh2015bayesian, brunskill2012bayes,guez2012efficient,lee2018bayesian,yu2017preparing}. Our work is similar to Bayesian-adaptive MDPs in terms of updating a belief distribution with Bayesian update rules, but we focus on the robustness against task estimate uncertainties at the same time.
The closest work to our research is \cite{xie2022robust}, which optimizes a conditional value-at-risk objective and maintains an uncertainty set centered on a context point estimate. Instead, we maintain an ambiguity set over beliefs and further consider the presence of task subpopulations.
\cite{sharma2019robust} also considers the uncertainties in belief estimates but with a flat latent task structure.

\vspace{-0.1in}
\paragraph{Multi-task RL.}
Learning a suite of tasks with an RL agent has been studied under different frameworks \cite{kirk2021survey, xu2022trustworthy}, such as Latent MDP \cite{kwon2021rl}, Multi-model MDP \cite{steimle2021multi}, Contextual MDP \cite{hallak2015contextualMD}, Hidden Parameter MDP \cite{doshi2016hidden}, and etc \cite{brunskill2013sample}. 
Our proposed HLMDP builds on the Latent MDP \cite{kwon2021rl} which contains a finite number of MDPs, each accompanied by a weight. In contrast to Latent MDP utilizing a flat structure to model each MDP's probability, HLMDP leverages a rich hierarchical model to cluster MDPs to a finite number of mixtures. In addition, HLMDP is a special yet important subclass of POMDP \cite{kaelbling1998planning}. It treats the latent task mixture that the current environment belongs to as the unobservable variable. HLMDP resembles the recently proposed Hierarchical Bayesian Bandit \cite{hong2021hierarchical} model but focuses on more complex MDP settings.


\section{Preliminary}
\label{sec:preliminary}
\vspace{-0.11in}
This section introduces Latent MDP and the adaptive belief setting, both serving as building blocks for our proposed HLMDP (Section~\ref{sec:HLMDP}) and GDR-MDP (Section~\ref{sec:GDR-MDP}).

\vspace{-0.1in}
\paragraph{Latent MDP.} An episodic Latent MDP \cite{kwon2021rl} is specified by a tuple $(\gM, T, \gS, \gA, \mu)$. 
$\gM$ is a set of MDPs with cardinality $|\gM|=M$.
Here $T$, $\gS$, and $\gA$ are the shared episode length (planning horizon), state, and action space, respectively. 
$\mu$ is a categorical distribution over MDPs and $\sum_{m=1}^M \mu(m) = 1$.
Each MDP $\gM_m \in \gM, m \in [M]$ is a tuple $(T, \gS, \gA, \ermP_m, \ermR_m, \nu_m)$ where $\ermP_m$ is the transition probability, $\ermR_m$ is the reward function and $\nu_m$ is the initial state distribution. 

Latent MDP assumes that at the beginning of each episode, one MDP from set $\gM$ is sampled based on $\mu(m)$. It aims to find a policy $\pi$ that maximizes the accumulated expected return solving $\max_{\pi} \sum_{m=1}^M \mu(m) \mathbb{E}_m^{\pi}\big[ \sum_{t=1}^T r_t \big]$, where $\mathbb{E}_m[\cdot]$ denotes $\mathbb{E}_{\ermP_m, \ermR_m}[\cdot]$.

\vspace{-0.1in}
\paragraph{The Adaptive Belief Setting} In general, a belief distribution contains the probability of each possible MDP that the current environment belongs to.
The adaptive belief setting \cite{steimle2021multi} holds a belief distribution that is dynamically updated with streamingly observed interactions and prior knowledge about the MDPs.
In practice, prior knowledge may be acquired by rule-based policies or data-driven learning methods. For example, it is possible to pre-train in simulated complete information scenarios or exploit unsupervised learning methods based on online collected data \cite{xu2020task}.
There also exist multiple choices for updating the belief, such as applying the Bayesian rule as in POMDPs \cite{kaelbling1998planning} and representing beliefs with deep recurrent neural nets \cite{karkus2017qmdp}.


\section{Hierarchical Latent MDP}
\vspace{-0.1in}
\label{sec:HLMDP}
In realistic settings, tasks share similarities, and task subpopulations are common. Although different MDP formulations are proposed to solve multi-task RL, the task relationships are in general overlooked. To fill in the gap, we first propose \textbf{Hierarchical Latent MDP (HLMDP)}, which utilizes a hierarchical mixture model to represent distributions over MDPs. Moreover, we consider the adaptive belief setting to leverage prior information about tasks.

\begin{definition}[Hierarchical Latent MDPs]\label{def:HLMDP}
    An episodic HLMDP is defined by a tuple $(\gZ, \gM, T, \gS, \gA, w)$.
    $\gZ$ denotes a set of Latent MDPs and $|\gZ|=Z$.
    $\gM$ is a set of MDPs with cardinality $|\gM|=M$ shared by different Latent MDPs.
    $T$, $\gS$, and $\gA$ are the shared episode length (planning horizon), state, and action space, respectively. 
    Each Latent MDP $\gZ_z \in \gZ, z \in [Z]$ consists of a set of joint MDPs $\{\gM_m\}_{m=1}^{M}$ and their weights $\mu_z$ satisfying $\sum_{m=1}^M \mu_z(m) = 1$. 
    $w$ is the categorical distribution over Latent MDPs and $\sum_{z=1}^Z w(z) = 1$. 
\end{definition}

\vspace{-0.05in}
We provide a graphical model of HLMDP in Figure~\ref{fig:illustration}~(c). 
HLMDP assumes that at the beginning of each episode, the environment first samples a Latent MDP $z \sim w(z)$ and then samples an MDP $m \sim \mu_z(m)$. 
HLMDP encodes task similarity information via the mixture model, and thus contains richer task information than Latent MDP proposed in \cite{kwon2021rl}. 
For instance, we could always find one Latent MDP for each HLMDP. However, there may exist infinitely many corresponding HLMDPs given one Latent MDP.

\vspace{-0.1in}
\paragraph{HLMDP in Adaptive Belief Setting.}
When solving multi-task RL problems, the adaptive setting is shown to help generate a policy with a higher performance \cite{steimle2021multi} than the non-adaptive one since it leverages prior knowledge about the transition model as well as the online collected data tailored to the unseen environment.
Hence we are motivated to formulate HLMDP in the adaptive belief setting. 

HLMDP maintains a belief distribution $b(z)$ over task groups to model the probability that the current environment belongs to each group $z$.
At the beginning of each episode, we initialize the belief distribution with a uniform distribution $b_0$. 
We use the Bayesian rule to update beliefs based on interactions and a prior knowledge base. Note that the knowledge base are not accurate enough and may lead to inaccurate belief updates. At timestep $t$, we get the next belief estimate $b_{t+1}$ with the state estimation function $SE$:
\begin{align}
    SE(b_t, s_t) =  \frac{b_t(j)L(j)}{\sum_{i \in [Z]} b_t(i)L(i)}, \forall j \in [Z],
    \label{eq:belief_update}
\end{align}
wher
Under the adaptive belief setting, HLMDP aims to find an optimal policy $\bar{\pi}^{\star}$ within a history-dependent policy class $\Pi$, under which the discounted expected cumulative reward is maximized as in Equation~\ref{eq:HLMDP}. Following general notations in POMDPs, we denote the history at time $t$ as $h_t = (s_0, a_1, s_1, \dots, s_{t-1}, a_{t-1}, s_t) \in \mathcal{H}_t$ containing state-action pairs $(s,a)$. At timestep $t$, we use both the observed state $s_t$ and the inferred belief distribution $b_t(z)$ as the sufficient statistics for history $h_t$.
\begin{align}
    \Bar{V}^{\star} &=\max_{\pi \in \Pi} \mathbb{E}_{b_{0:T}(z)} \mathbb{E}_{\mu_z(m)} \mathbb{E}_m^{ \pi} \big[\sum_{t=1}^{T} \gamma^t r_t \big], \label{eq:HLMDP}
\end{align}
where $r_t$ denotes the reward received at step $t$. $b_0(z)$ is the initial belief at timestep 0.


\section{Group Distributionally Robust MDP}
\label{sec:GDR-MDP}
\vspace{-0.1in}

The belief update function in Equation~\ref{eq:belief_update} may not be accurate, which motivates robust decision-making under belief estimate errors. In this section, we introduce \textbf{Group Distributionally Robust MDP (GDR-MDP)} which models task groups and considers robustness against the belief ambiguity.
We then study the convergence property of GDR-MDP in the infinite-horizon case in Section~\ref{sec:convergence}.
We find that GDR-MDP's hierarchical structure helps restrict the worst-possible value within the ambiguity set and provide the robustness guarantee in Section~\ref{sec:analysis}.

\begin{definition}[General Ambiguity Sets]
Let $\Delta^k$ be a $k$-simplex. 
Considering a categorical belief distribution $b \in \Delta^k$, a general ambiguity set without special structures is defined as $\gC_{\Delta^k}$ containing all possible distributions for $b$.
\label{def:general_ambiguity_set}
\end{definition}

\begin{definition}[Group Distributionally Robust MDP]\label{def:GDR-MDP}
An episodic GDR-MDP is defined by a 8-tuple $(\gC, \gZ, \gM, T, \gS, \gA, w, SE)$. $\gC$ is a general belief ambiguity set. $T, \gS, \gA, \gM, \gZ, w$ are elements of an episodic HLMDP as in Definition~\ref{def:HLMDP}. $SE: \Delta^{Z-1} \times \gS \rightarrow \Delta^{Z-1}$ is the belief updating rule.
GDR-MDP aims to find a policy $\pi^\star\in \Pi$ that obtains the following optimal value:
\begin{align}
    V^{\star} &=\max_{\pi \in \Pi} \min_{ \substack{ \hat{b}_{0:T} \\ \in \gC_{\Delta^{Z-1}} } } \mathbb{E}_{\hat{b}_{0:T}(z)} \mathbb{E}_{\mu_z(m)} \mathbb{E}_m^{ \pi} \big[\sum_{t=1}^{T} \gamma^t r_t \big], \label{eq:GDRO} 
\end{align}
where $\gC_{\Delta^{Z-1}}$ is a general ambiguity set tailored to beliefs over Latent MDPs in set $\gZ$. 
\end{definition}

\vspace{-0.05in}
GDR-MDP naturally balances robustness and performance by leveraging distributionally robust formulation and rich distributional information.
In contrast to HLMDP, which maximizes expected return over nominal adaptive belief distribution (Equation~\ref{eq:HLMDP}), GDR-MDP aims to maximize the expected return under the worst-possible beliefs within an ambiguity set $\gC_{\Delta^{Z-1}}$.
Moreover, GDR-MDP optimizes over fewer optimization variables than when directly perturbing MDP model parameters or states. 
It resembles the group distributionally robust optimization problem in supervised learning \cite{sagawa2019distributionally, oren2019distributionally} but focuses on sequential decision-making in dynamic environments.

\vspace{-0.1in}
\subsection{Convergence in Infinite-horizon Case}
\vspace{-0.1in}
\label{sec:convergence}

With general ambiguity sets (as in Definition~\ref{def:general_ambiguity_set}), calculating the optimal policy is intractable \cite{yu2015distributionally, Delage2010DistributionallyRO}.
We propose a belief-wise ambiguity set that follows the b-rectangularity to facilitate solving the proposed GDR-MDP.

\vspace{0.05in}
\begin{assumption}[b-rectangularity]
We assume a belief-wise ambiguity set, $\Tilde{\gC}:= \bigotimes_{b \in \Delta^{Z-1}} \gC_b$, where $\bigotimes$ represents Cartesian product.
$b$ serves as the nominal distribution of the ambiguity set.
\label{assumption:b_rec}
\end{assumption}

\vspace{-0.05in}
More concretely, the b-rectangularity assumption uncouples the ambiguity set related to different beliefs.
When conditioned on beliefs at each timestep, the minimization loop selects the worst-case realization unrelated to other timesteps. 
The b-rectangularity assumption  is motivated by the $s$-rectangularity first introduced in \cite{wiesemann2013robust}, which helps reduce a robust MDP formulation to an MDP formulation and get rid of the time-inconsistency problem \cite{xin2021time}. 
Ambiguity sets beyond rectangularities are recently explored in \cite{mannor2016robust,goyal2018robust}, which we leave for future works.

With b-rectangular ambiguity sets, we derive Bellman equations to solve Equation~\ref{eq:GDRO} with dynamic programming. Detailed proofs are in Appendix Section~\ref{sec:proof_convergence}.

\vspace{0.05in}

\begin{proposition}[Group Distributionally Robust Bellman Equation]\label{prop_bellman}
Define the distributionally robust value of an arbitrary policy $\pi$ as follows where $b_{t+1} = SE(b_t, s_t)$.
\begin{align*}
    V_t^{\pi}(b_t, s_t) \!  =  \! \min_{ \substack{ \hat{b}_{t:T} \in \\ \gC_{b_{t:T}} } }  \mathbb{E}_{\hat{b}_{t:T}(z)} \mathbb{E}_{\mu_{z}(m)} \mathbb{E}_{m}^{\pi_{t:T}} \! \big[  
     \sum_{n=t}^{T} \gamma^{n-t} r_n  \vert b_t, s_t \big].
\end{align*}
The Group Distirbutionally Robust Bellman expectation equation is
\begin{align}
    V_t^{\pi}(b_t, s_t) &= \min_{\hat{b}_{t} \in \gC_{b_{t}}} \mathbb{E}_{\hat{b}_{t}(z)} \mathbb{E}_{\mu_z(m)} \mathbb{E}^{\pi_t} \Big[ \mathbb{E}_{\ermR_{m}}[r_t] + \nonumber \\
    &\gamma \sum_{s_{t+1}} \ermP_m(s_{t+1}| s_t, a_t) V^{\pi}_{t+1}(b_{t+1}, s_{t+1})   \Big]. \label{eq:bellman_exp}
\end{align} 
\end{proposition}

\begin{lemma} [Contraction Mapping] Let $\mathcal{V}$ be a set of real-valued bounded functions on $\Delta^{Z-1} \times \mathcal{S}$. $\mathcal{L}V(b, s): \mathcal{V} \rightarrow \mathcal{V}$ refers to the Bellman operator defined as 
\begin{align}
    \mathcal{L}V(b, s) &=\max_{\pi \in \Pi} \min_{\hat{b} \in \gC_{b}} \mathbb{E}_{\hat{b}(z)} \mathbb{E}_{\mu_z(m)} \mathbb{E}^{\pi} \Big[ \mathbb{E}_{\ermR_{m}}[r] + \nonumber \\
    &\gamma \sum_{s'} \ermP_m(s'| s, a) V^{\pi}(SE(b, s), s)   \Big]. \label{eq:contraction}
\end{align}
$\mathcal{L}V(b,s)$ is a $\gamma$-contraction operator on the complete metric space $(\mathcal{V}, \| \cdot \|_{\infty})$. That is, given $ \forall \ U, V \in \mathcal{V}$, $\| \mathcal{L}U - \mathcal{L}V \|_{\infty} \leq \gamma \| U-V \|_{\infty}$. 
\label{lemma:contraction}
\end{lemma}

\vspace{0.05in}
\begin{theorem}[Convergence in Infinite-horizon Case]
    Define $V_{\infty}(b, s)$ as the infinite horizon value function.
    For all $b \in \mathcal{B}$ and $s \in \mathcal{S}$, we have $V_{\infty}(b, s)$ is the unique solution to $\mathcal{L}V_{\infty}(b, s) = V_{\infty}(b, s)$, and $ \lim_{t\rightarrow \infty}\mathcal{L}V_{t}(b, s) = \mathcal{L}V_{\infty}(b, s)$ uniformly in $\|\cdot \|_{\infty}$.
    \label{theorem_infinite_convergence}
\end{theorem}
By repeatedly applying the contraction operator in Lemma~\ref{lemma:contraction}, the value function will converge to a unique fixed point, which corresponds to the optimal value based on Banach fixed point theorem \cite{banach1922operations}.

\vspace{-0.1in}
\subsection{Robustness Guarantee of GDR-MDP}
\vspace{-0.1in}
\label{sec:analysis}

This section shows how GDR-MDP's hierarchical task structure and the distributionally robust formulation help balance performance and robustness. 
We compare the optimal value of GDR-MDP denoted as $V_{GDR}(\pi^{\star}_{GDR})$, with three different robust formulations.
Group Robust MDP is a robust version of GDR-MDP with its optimal value denoted as $V_{GR}(\pi^{\star}_{GR})$.
Distributionally Robust MDP holds a belief over MDPs without the hierarchical task structure whose optimal value denoted as $V_{DR}(\pi^{\star}_{DR})$.
 Robust MDP is a robust version of Distributionally Robust MDP, denoted as $V_{R}(\pi^{\star}_{R})$.
$\pi^{\star}_{\cdot}$ denote optimal policies under different formulations. 
We achieve the comparison by studying how maintaining beliefs over mixtures affects the worst-possible outcome of the inner minimization problem and the resulting RL policy. 

\vspace{-0.05in}
We study the worst-possible value via the relationships between ambiguity sets projected to the space of beliefs over MDPs.
We first define a discrepancy-based ambiguity set that is widely used in existing DRO formulations \cite{abdullah2019wasserstein, sinha2017certifying, lecarpentier2019non}. 
\begin{definition}[Ambiguity set with total variance distance]
\label{def:tv_set} Consider a discrepancy-based ambiguity set defined based on total variance distance. Formally, the ambiguity set is
\begin{align*}
    \gC_{\nu_{\gX}, d_{TV}, \xi}(X) = \{ \nu'(X) : \sup_{X \in \gX}|\nu'(X) - \nu_{\gX}(X)| \leq \xi\},
\end{align*}
where $X \in \gX$ is the support, $\nu_{\gX}$ is the nominal distribution over $\gX$ and $\xi$ is the ambiguity set's size.
\end{definition}

To achieve a reasonable comparison, we control the adversary's budget $\xi$ the same when perturbing the belief over task groups $z$ and tasks $m$, which correspond to different model misspecification forms when there is a hierarchical latent structure about tasks.

\begin{theorem}[Values of different robust formulations]
Let
$U_{m}(\pi) = \mathbb{E}_m^{ \pi} \big[\sum_{t=1}^{T} \gamma^t r_t \big]$.
Let $\gC_{b(m), d_{TV}, \xi}(m)$ and $\gC_{b(z), d_{TV}, \xi}(z)$ denote the ambiguity sets for beliefs over tasks $m$ and groups $z$, respectively. $b(m)$ and $b(z)$ satisfy $b(m) = \sum_{\gZ}\mu_z(m)b(z)$ and are the nominal distributions.
For any history-dependent policy $\pi \in \Pi$, its value function under different robust formulations are:
\begin{align*}
    V_{GDR}(\pi) &=  \min_{\hat{b}(z) \in \gC_{b(z), d_{TV}, \xi}(z)} \mathbb{E}_{\hat{b}(z)} \mathbb{E}_{\mu_z(m)}  [ U_{m}(\pi) ], 
    \\
    V_{GR}(\pi) &= \min_{z \in [Z]} \mathbb{E}_{\mu_z(m)}  [ U_{m}(\pi) ], \\
    V_{DR}(\pi) &= \min_{\hat{b}(m) \in \gC_{b(m), d_{TV}, \xi}(m)} \mathbb{E}_{\hat{b}(m)}  [ U_{m}(\pi) ],
    \\
    V_{R}(\pi) &=  \min_{m \in [M]}[ U_{m}(\pi) ].
\end{align*}

We have the following inequalities hold: $V_{GDR}(\pi) \geq V_{GR}(\pi) \geq V_{R}(\pi)$ and $V_{GDR}(\pi) \geq V_{DR}(\pi)$.

\label{theorem:value_analysis}
\end{theorem}

\vspace{-0.1in}
Theorem~\ref{theorem:value_analysis} shows that with a nontrivial ambiguity set, the distributionally robust formulation in GDR-MDP helps regularize the worst-possible value when compared with robust ones, including the group robust (GR) and task robust (R) formulations. 
It also shows that GDR-MDP's hierarchical structure further helps restrict the effect of the adversary, resulting in higher values than the distributionally robust formulation with a flat latent structure (DR).
To get Theorem~\ref{theorem:value_analysis}, we first find that when projecting the $\xi$-ambiguity set for $b(z)$ to the space of $b(m)$, the resulting ambiguity set is a subset of the $\xi$-ambiguity set for $b(m)$. Proofs are detailed in Appendix Section~\ref{sec:appendix_robustness}.
Our setting is different from \cite{staib2017distributionally} which states that DRO is a generalization of point-wise attacks. 
The key difference is that when the adversary perturbs $b(m)$, we omit the expectation over the mixtures under $b(z)$.

\begin{theorem}[Optimal values of different robust formulations]
Let $\pi_{\cdot}^{\star}$ denote the converged optimal policy for different robust formulations, we have
$V_{GDR}(\pi_{GDR}^{\star}) \geq V_{GR}(\pi_{GR}^{\star}) \geq V_{R}(\pi_{R}^{\star}) $ and $V_{GDR}(\pi_{GDR}^{\star}) \geq V_{DR}(\pi_{DR}^{\star}) $.
\label{theorem:optimal_policies}
\end{theorem}

\vspace{-0.05in}
Based on Theorem~\ref{theorem:value_analysis}, we can compare the optimal values for different robust formulations. Theorem~\ref{theorem:optimal_policies} shows that imposing ambiguity set on beliefs over mixtures helps generate less conservative policies with higher optimal values at convergence compared with other robust formulations.

\begin{figure}[t]
    \centering
    \includegraphics[width=0.95\linewidth]{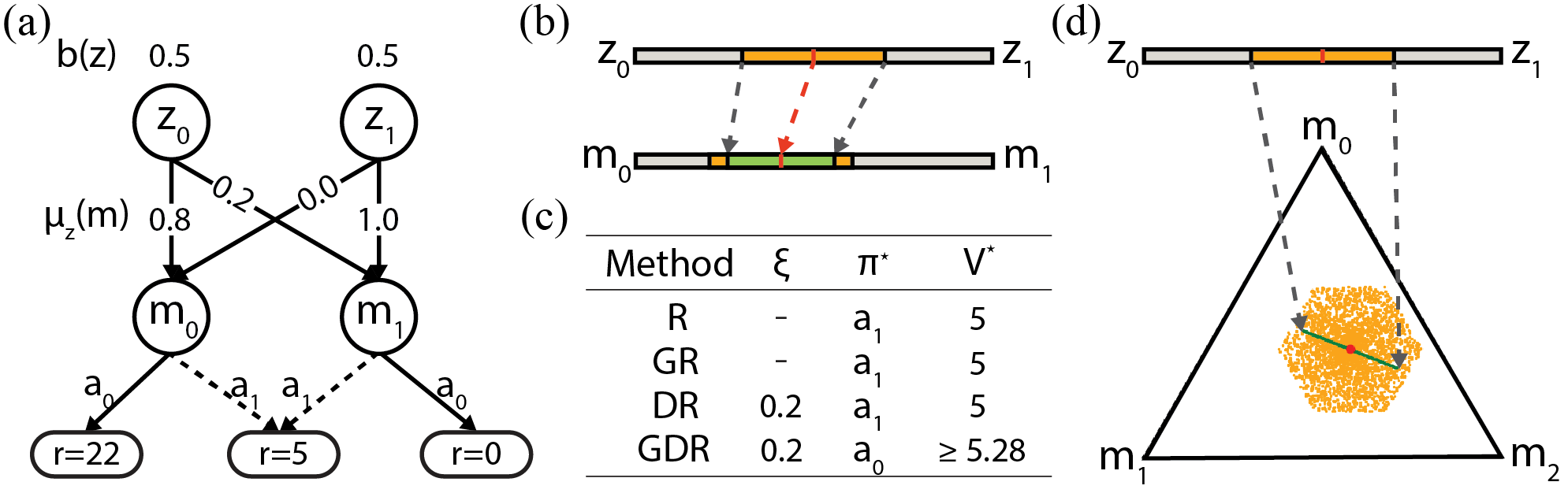}
    \vspace{-0.1in}
    \caption{Hierarchical Latent Bandit examples. 
    (a), (b) and (c) show the graphical model, the relationship between ambiguity sets, and different robust formulations' optimal values for an example with two groups and two unique tasks. 
    (d) shows the relationship between ambiguity sets for an example with two groups and three unique tasks. 
    }
    \label{fig:toy_example}
\end{figure}

\vspace{-0.05in}

\textbf{Illustration Examples in Figure~\ref{fig:toy_example}.} We provide two hierarchical latent bandit examples in Figure~\ref{fig:toy_example}. 
The first example shown in Figure~\ref{fig:toy_example}~(a) has two latent groups with different weights over two unique MDPs. 
(b) shows the ambiguity sets of the example in (a). The orange sets denote the $\xi$-ambiguity sets for the beliefs over mixtures and MDPs. The green set denotes the ambiguity set projected from the $\xi$-ambiguity set for belief distributions over mixtures. We show that the mapped set is a subset of the original $\xi$-ambiguity set for the MDP belief distributions. 
(c) shows the optimal policy and value of different robust formulations for the example in (a). Our proposed GDR has the potential to get a less conservative policy with higher returns than other robust baselines.
(d) follows the same notations in (b) but corresponds to an example with three possible MDPs. (b) and (d) together shows that the hierarchical structure helps regularize the adversary's strength. The detailed procedure for getting the optimal policies is shown in Appendix~\ref{sec:appendix_toy_example}.

\vspace{-0.1in}
\section{Algorithms}
\label{sec:algo_implement}
\vspace{-0.1in}

To solve the proposed GDR-MDP, we propose novel robust deep RL algorithms (summarized in Algorithm~\ref{alg:GDR_DQN_GDR_SAC} and Algorithm~\ref{alg:GDR_PPO} in appendix), including GDR-DQN based on Deep Q learning \cite{mnih2013playing}, GDR-SAC based on soft actor-critic \cite{haarnoja2018soft}, and GDR-PPO based on PPO~\cite{schulman2017proximal}. 
We learn robust policies that take the inferred belief distribution over mixtures $b(z)$ and the state $s$ as input.
We implement GDR-DQN and GDR-SAC with Tianshou \cite{weng2021tianshou} and GDR-PPO with stable-baselines3 \cite{stable-baselines3}. Details are in Appendix Section~\ref{sec:appendix_implementation_detail}.

\begin{algorithm}[t]
\SetAlgoLined
\SetNoFillComment
\caption{GDR-MDP Trajectory Rollout }
\label{alg:rollout}
    \KwIn{Mixing weights $w(z)$ and $\mu_z(m)$, episode index $n$, episode length $T$, belief update function $SE$, rollout policy $\pi_{\theta}(b(z), s)$, exploration $\epsilon$} 
    {\bfseries Initialize} episodic history $h=\{ \}$ \;
    Sample mixture $z_n \sim w(z)$ \;
    Sample MDP $m_n \sim \mu_{z_n}(m)$ \;
    Initialize belief $b_0(z)$ as a uniform distribution \;
    \For{$t=0$ {\bfseries to} $T$}{
    Sample action $a_t$ with the $\epsilon$-greedy method and rollout in MDP $m$\;
    $b_{t+1}(z) = SE(b_t(z), s_{t+1})$ \;
    Append the most recent data pair $d=\{ (b_{t}, s_t), a_t, r_t, (b_{t+1},s_{t+1}) \}$ to $h$ \;}
    {\bfseries Return:} history $h$, episode return
\end{algorithm}

\vspace{-0.1in}
\paragraph{GDR-DQN and GDR-SAC.} 
We update the Q-net in GDR-DQN and the critic net in GDR-SAC toward TD targets with perturbed beliefs.
We follow Definition~\ref{def:tv_set} to construct the ambiguity set $\gC_{b'(z), d_{TV}, \xi}$ which centers at the originally inferred $b'(z)$ and satisfies the b-rectangularity assumption stated in Assumption~\ref{assumption:b_rec}.
At each training step, we sample a batch data $\{ d=(b(z), s, a, r, b'(z), s', a', r' ) \}^N$ from the replay buffer $\gD$ to estimate the perturbed TD target.

We update Q-functions with gradient descents. For both GDR-DQN and GDR-SAC, we have loss as 
\begin{align*}
    \gL_{Q_{\theta}} = \sum_{  \substack{d } }  \Big( &r + \min_{ \substack{p(z) \in \\  \gC_{b'(z), d_{TV}, \xi}}  } \sum_{a\in \gA} Q_{\theta} ( p(z) , s', a) \\
    &- \sum_{a\in \gA} Q_{\theta} (b(z),s,a) \Big)^2.
\end{align*}

\vspace{-0.2in}
\paragraph{GDR-PPO.} GDR-PPO conducts robust training by decreasing the advantages of trajectories that are vulnerable to belief noises. More concretely, given a trajectory $d$, its advantage for $(b_t, s_t)$ is calculated as follows.
\begin{align}
    \hat{A}(b_t, s_t) &= \sum_{t'=t}^{T-1} r_t - R_{drop} - V_{\theta}(b_t, s_t), \text{ where} \nonumber \\
    R_{drop} &=  V(b_t, s_t) - \min_{ \substack{p(z) \in   \gC_{b_t(z), d_{TV}, \xi}}  } V_{\theta} ( p(z) , s_t) . \nonumber
\end{align}
We measure the performance drop $R_{drop}$ under worst-possible beliefs within the ambiguity set.

\vspace{-0.1in}
\paragraph{Worst-possible Beliefs.}
To obtain the worst case distribution $b^{adv}  \in \gC_{b'(z), d_{TV}, \xi}$, we iteratively apply a stochastic variant of fast gradient sign method (FGSM) \cite{goodfellow2014explaining} to make sure that the perturbed discrete distribution satisfies $\sum_{z}p(z) = 1$.
For each attack to the belief distribution, we randomly sample an index $i \in Z$, and apply the attack to each element in $p(z)$ as follows $p(z)_j = p(z)_j + \alpha_{b} \cdot \sign ( \nabla_{p(z)_j} V(p(z), s')), \forall j\not=i$ and $p(z)_i = p(z)_i - \sum_{j \not=i} p(z)_j$. 
$\alpha_{b}$ is the perturbation step size. 
To stabilize robust training, we pretrain for a small amount of episodes with exact one-hot beliefs to ensure that the value function could approximate the actual state value to some extent.
To achieve a certain level of robustness over noisy inferred belief $b(z)$, we fix the ambiguity set size along with robust training, which is analogous to the adversary budget and the robustness level \cite{zhou2021finite}.

\begin{figure*}[t]
    \vspace{-0.15in}
    \begin{center}
        \includegraphics[width=0.9\linewidth]{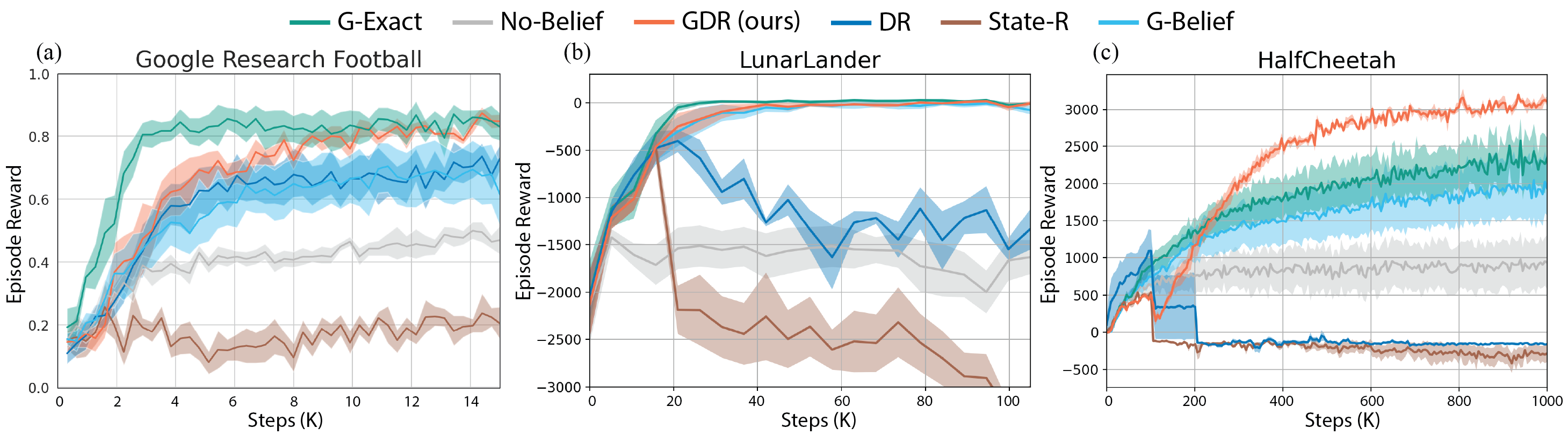}
    \end{center}
     \vspace{-0.2in}
    \caption{The training performance of GDR and baselines. 
    Each curve is averaged over 5 runs and shaded areas represent standard errors. Our results show that GDR has better training stability than DR by implicitly regularizing the adversary's strength with the hierarchical structure.
    }
    \label{fig:training}
    \vspace{-0.05in}
\end{figure*}

\begin{algorithm}[t]
\SetAlgoLined
\SetNoFillComment
\caption{Group Distributionally Robust Training for GDR-DQN and GDR-SAC
}
\label{alg:GDR_DQN_GDR_SAC}
    \KwIn{Q-net $Q_{\theta}(b(z), s, a)$, ambiguity set $\gC_{\cdot, d_{TV}, \xi}$, training episodes $N$, }
    {\bfseries Initialize} data buffer $\gD$ \;
    \For{$n=0$ {\bfseries to} $N$}{
     Rollout one episode with Algorithm~\ref{alg:rollout} and append data pairs to $\gD$ \;
    \If{Update Q-net parameters}{
     Sample batch data from $\gD$ \;
    \For{Each $d_i$ in the batch}{
        Get $b^{adv} \in \gC_{b'(z), d_{TV}, \xi}$ with modified FGSM\;
    }
    Update Q-net $\theta \leftarrow \theta - \alpha_{\theta} \nabla_{\theta}\gL_{Q_{\theta}}$\;
    }
    }    Return: Q-net $Q_{\theta}$
\end{algorithm}

\vspace{-0.1in}
\section{Experiments}
\label{sec:experiments}
\vspace{-0.1in}
We conduct experiments to empirically study (a) the effect of GDR-MDP's hierarchical structure on the robust training stability and (b) policy's robustness to belief estimate error.

\vspace{-0.1in}
\subsection{Environments}
\vspace{-0.12in}
We evaluate GDR-DQN in Lunarlander \cite{brockman2016openai}, GDR-SAC in Halfcheetah \cite{todorov2012mujoco}, and GDR-PPO in Google Research Football \cite{kurach2020google}. Table~\ref{tab:environments} shows a summary of environment setups. More details are in Appendix Section~\ref{sec:appendix_environments}. 
To initialize each episode, we first sample a group $z \sim w(z)$, and then a task $m \sim \mu_z(m)$ for the episode. Note that both $z$ and $m$ are unknown to the agent. 

\vspace{-0.15in}
\paragraph{Google Research Football (GRF).}
This domain presents additional challenges due to its AI randomness, large state-action spaces, and sparse rewards. 
The RL agent will control one active player on the attacking team at each step and can pass to switch control. The non-active players will be controlled by built-in AI. The dynamics of our designed 3 vs. 2 tasks are determined by the player types including central midfield (CM) and centre back (CB), and player capability levels. 
The built-in CM player tends to go into the penalty area when attacking and guard the player on the wing (physically left or right) when defending, while the CB player tends to guard the player in the middle when defending, and not directly go into the penalty area when attacking. 
Different patterns of policies are required to solve the tasks from different groups.

\vspace{-0.15in}
\paragraph{Box2D Control Task: LunarLander.}
The Lunarlander's dynamics are controlled by the engine mode and engine power. In the flipped mode, the action turning on the left (or right) engine in normal mode will turn on the right (or left) engine instead. 

\vspace{-0.15in}
\paragraph{Mujoco Control Task: HalfCheetah.}
In HalfCheetah, each task's dynamics are controlled by both the torso mass and the failure joint, to which we cannot apply action. Our setting is similar to the implementation in \cite{xie2022robust} but with a fixed failure joint within each episode.

\vspace{-0.1in}
\subsection{Baselines}
\vspace{-0.12in}

\begin{table}[t]
\vspace{0.15in}
    \centering
    \caption{Environment setups. Both parameters affect the environment dynamics.
    In GRF, the strongest player has a capability level of 1.0.
    Our tasks are more challenging than the original 3 vs. 2 task in GRF (1.0 vs. 0.6) in terms of the capability level.
    For notation simplicity, let $\mathbf{1}^k$ be a k-dimensional vectored filled with 1, and $E(k) = \frac{1}{k}\mathbf{1}^k$. $E_i(k)$ shows the $i$-th matrix block on the diagonal. }
    \vspace{-0.05in}
    \resizebox{0.5\textwidth}{!}{
    \begin{tabular}{cccc}
    \toprule
        Environment & GRF (3 vs. 2) & LunarLander & HalfCheetah \\ \midrule
        Parameter 1  &  Player Type  &  Engine Mode & Failure Joint \\
        (Mixture) & \{CM vs. CB, CB vs. CM\}   & \{Normal, Flipped\} & \{0,1,2,3,4,5\} \\ \midrule
        \multirow{2}{*}{Parameter 2} & Player Capability Level & Engine Power & Torso Mass\\
        & \{0.9 vs. 0.6, 1.0 vs. 0.7\} & \{3.0, 6.0\} & \{0.9, 1.0, 1.1\} \\ \midrule
        \# Mixtures & 2 & 2 & 6 \\ \midrule
        $w$ & [0.5, 0.5] & [0.5, 0.5] & $\frac{1}{6} \mathbf{1}^6$ \\ \midrule
        \# MDPs & 4 & 4 & 18 \\ \midrule
        $\mu_z(m)$ & 
        $
        \begingroup
        \setlength\arraycolsep{1pt}
        \begin{bmatrix} \frac{1}{2}\mathbf{1}^2 & \mathbf{0} \\ \mathbf{0} & \frac{1}{2}\mathbf{1}^2 \end{bmatrix}
        \endgroup
        $ &
        $
        \begingroup
        \setlength\arraycolsep{0.0001pt}
        \begin{bmatrix} \frac{1}{2}\mathbf{1}^2 & \mathbf{0} \\ \mathbf{0} & \frac{1}{2}\mathbf{1}^2 \end{bmatrix}
        \endgroup
        $ &
        $
        \begingroup
        \setlength\arraycolsep{1pt}
        \begin{bmatrix} 
        E_0(6) & \mathbf{0} & \mathbf{0} \\ 
        \mathbf{0} & \cdots  & \mathbf{0} \\ 
        \mathbf{0} & \mathbf{0} & E_
        5(6)\\ 
        \end{bmatrix} 
        \endgroup
        $ 
        \\ 
        \bottomrule
    \end{tabular}
    }
    \label{tab:environments}
    \vspace{-0.1in}
\end{table}

We compare our Group Distributionlly Robust training methods (\textbf{GDR}) with five baselines. 
In \textbf{G-Exact}, the RL agent is trained with the exact mixture information encoded in a one-hot vector.
The agent in \textbf{DR} maintains a belief distribution $b(m)$ and utilizes distributionally robust training over $b(m)$. It uses the same belief updating rule as in GDR to update $b(z)$ at each timestep but projects $b(z)$ to $b(m)$ with $\mu_z(m)$. DR utilizes no mixture information and helps ablate the effect of the hierarchical latent structure. 
The agent in \textbf{No-Belief} has no access to the context information and generates action only based on state $s$. The No-Belief baseline helps show the importance of the adaptive belief setting.
In \textbf{G-Belief}, the agent maintains belief $b(z)$ and is trained towards a nominal TD target. Compared with GDR, G-Belief helps reveal the effect of distributionally robust training.
The \textbf{State-R} agent takes both the inferred belief $b(z)$ and state $s$ as input. It updates towards a TD target with perturbed states along with training. For baselines with belief modules, we utilize the Bayesian update rule in Equation~\ref{eq:belief_update} and leave the detailed likelihood calculation in Appendix Section~\ref{sec:appendix_implementation_detail}.

\vspace{-0.1in}
\section{Results and Discussion}
\label{sec:discussion}
\vspace{-0.1in}

\begin{figure*}
    \vspace{-0.1in}
    \begin{center}
        \includegraphics[width=0.9\linewidth]{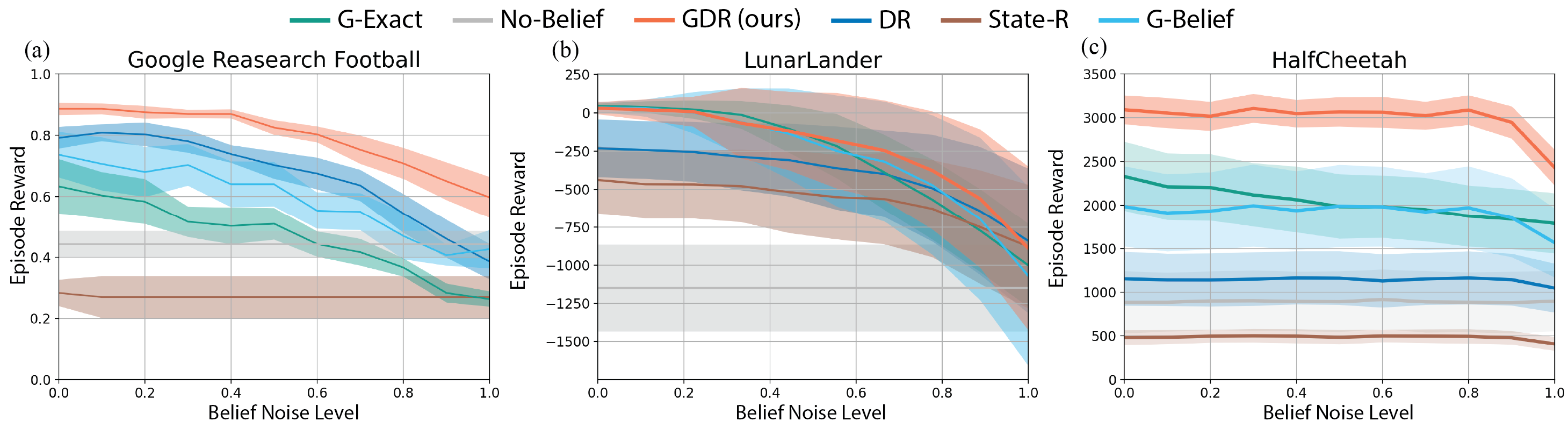}
    \end{center}
     \vspace{-0.2in}
    \caption{Robustness evaluations when facing belief inference errors.
    Each plot is averaged over 5 runs and shaded areas represent standard errors. GDR preserves higher robustness to belief inference errors compared with baselines.
    }
    \label{fig:robustness}
    \vspace{-0.05in}
\end{figure*}

\subsection{Influence of the GDR-MDP's Hierarchical Structure on Robust Training}
\vspace{-0.1in}
We study the effect of the hierarchical structure on the adversary's strength based on training performances in Figure~\ref{fig:training}. We show the importance of mixture information since the No-Belief baseline consistently underperforms G-Exact during training in all three environments. Lunarlander and HalfCheetah have a return much lower than G-Exact since the kinematic observation fed into the neural net does not reveal any mixture information.
In GRF, the No-Belief baseline underperforms G-Exact since it could not effectively learn distinct strategies with regard to different types of players as teammates and opponents, while G-Exact could learn group-specific policies.

\vspace{-0.05in}
When compared with other robust training baselines including DR and State-R,
GDR achieves a higher average return at convergence in all environments as in Figure~\ref{fig:training}. In LunarLander and HalfCheetah, DR which maintains a belief $b(m)$ over MDPs induces significant training instability, instead of learning a meaningful conservative policy.
In GRF, DR has a worse asymptotic performance than GDR.
Those observations empirically validate our theoretical result (Section~\ref{sec:analysis}) in the regime of deep RL, which is that, with the same ambiguity set size, perturbing $b(m)$ omitting mixture information will lead to larger value perturbations than perturbing $b(z)$ over mixtures. The State-R baseline leads to more considerable training instability than DR and fails to learn in all three environments. 

\vspace{-0.05in}
We compare GDR with non-robust training baselines, including G-Exact and G-belief to study the importance of robust training.
In LunarLander, GDR has comparable training performance with G-Exact and G-Belief.
In GRF, GDR has slightly worse asymptotic performance than G-Exact and better performance than G-Belief.
These observations show that GDR successfully extracts task-specific information stored in the noisy beliefs and conditions on the beliefs for action generation. 
In HalfCheetah, GDR performs better than G-Exact. Although GDR leads to an immediate performance drop after pretraining (100000 steps), the robust training in GDR converges to higher performance. We conjecture that this is due to the perturbed belief helping the algorithm get out of local optima.

\vspace{-0.15in}
\subsection{Robustness to Belief Inference Errors}
\vspace{-0.15in}
We test the robustness against belief noise of the best policies obtained with GDR and baselines along with training. The results are shown in Figure~\ref{fig:robustness}.
We define the belief noise level as the inaccuracy of the likelihood when updating belief with the Bayesian rule. During robustness evaluation, G-Exact generate actions conditioned on the same noisy beliefs as GDR and G-Belief.

\vspace{-0.05in}
In GRF and HalfCheetah, GDR is consistently more robust to belief noise than robust and nominal training baselines.
In LunarLander, the mean reward of GDR is better than G-Exact when there is a high belief noise level and is better than DR when a low belief noise level.
The large variances in LunarLander are due to the large penalty when crashes which are further exaggerated by the fixed episode length. 
Although GDR has its performance decreasing along with the increase of the belief noise level, its performance is still an upper bound of DR and G-Exact's performances. 
These observations show that GDR successfully balances the information between belief distributions and states, and is more robust to belief inference errors.

\vspace{-0.05in}
G-Exact is prone to injected belief noise since it heavily relies on accurate mixture information to achieve high performance.
G-Belief does not show significant robustness improvement over G-Exact. It shows that the group distributionally robust training procedure instead of the belief randomness along training helps improve the robustness.

\begin{figure}
    \vspace{0.05in}
    \centering
    \includegraphics[width=0.9\linewidth]{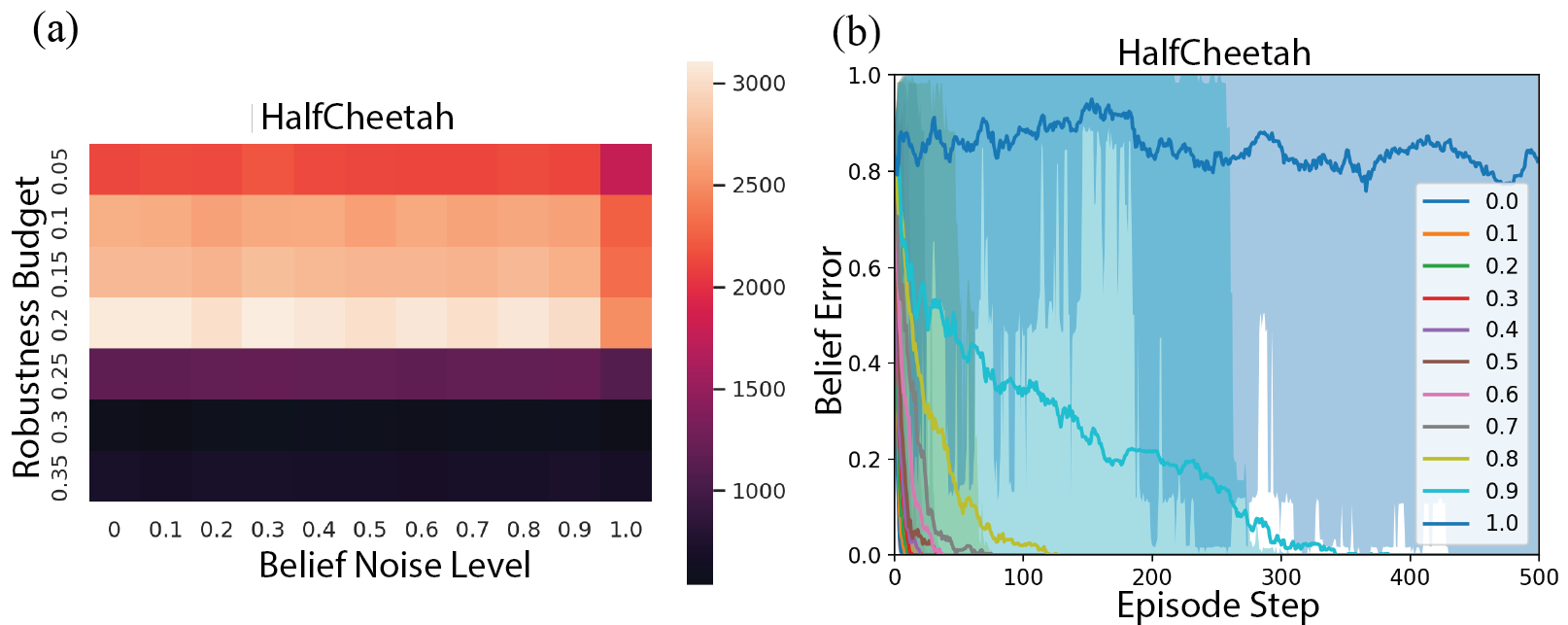}
    \vspace{-0.1in}
    \caption{Ablation studies in HalfCheetah.}
    \vspace{-0.1in}
    \label{fig:ablation}
\end{figure}

\vspace{-0.1in}
\subsection{Ablation Study}
\vspace{-0.1in}
We perform empirical sensitivity analysis to reveal the effect of uncertainty set size on GDR's policy robustness in HalfCheetah. Figure~\ref{fig:ablation} (a) shows that gradually increasing the ambiguity set size up to 0.2 helps improve the robustness. The ambiguity set whose size is greater or equal to 0.25, easily leads to training instability and thus decreases the robustness. 
In contrast, even with an ambiguity set of size 0.05 and pretraining for 300000 steps, DR without the mixture information still causes unstable training (see Appendix Section~\ref{sec:appendix_ablation}). 
Figure~\ref{fig:ablation} (b) provides the average belief errors at each time step corresponding to different belief noise levels.
Figure~\ref{fig:ablation} (b) and Figure~\ref{fig:robustness} show that GDR only shows significant performance drops when the belief error is nonzero for a large portion of steps.

\vspace{-0.12in}
\section{Conclusion}
\vspace{-0.13in}

This paper considers robustness against task estimate uncertainties. We propose the GDR-MDP formulation that can leverage rich distribution information, including adaptive beliefs and prior knowledge about task groups. To the best of our knowledge, GDR-MDP is the first distributionally robust MDP formulation that models ambiguity over belief estimates in an adaptive setting. We theoretically show that GDR-MDP's hierarchical latent structure helps enhance its distributional robustness compared with a flat task structure. 
We also empirically show that our proposed group distributionally robust training methods generate more robust policies than baselines when facing belief inference errors in realistic scenarios. 
We hope this work will inspire future research on how diverse domain knowledge affects robustness and generalization. 
One exciting future direction is to scale the group distributionally robust training to high-dimensional and continuous latent task distributions for diverse decision-making applications.

%% file: contents/appendix.tex


\setlength{\tabcolsep}{3pt}

\section{Toy Example: Hierarchical Latent Bandit}
\label{sec:appendix_toy_example}
\vspace{-0.1in}
In this section, we show the process of getting the optimal policies for different robust formulations in the Hierarchical Latent Bandit problem as illustrated in Figure~\ref{fig:toy_example} (a). 

The agent has two possible actions, $a_0$ and $a_1$. There are two possible mixtures/groups denoted as $z$, and two possible MDPs denoted as $m$. Given the mixture, we have the conditional probability for each MDP as $\mu(m=0|z=0)=0.8$, $\mu(m=1|z=0)=0.2$, $\mu(m=0|z=1)=0.0$, $\mu(m=1|z=1)=1.0$. We assume the same type of ambiguity set measured by the total variance distance as in the analysis. Let the current belief over groups be $b(z) = [0.5, 0.5]$ and the ambiguity set size be $\xi=0.2$.

We compare the optimal policies of four robust formulations, including 
\begin{itemize}
    \item our proposed GDR-MDP (shorthanded as \textbf{GDR}) that utilizes both the hierarchical structure and distributionally robust formulation, and optimizes over the worst-possible beliefs over groups,
    \item group robust MDP (\textbf{GR}), which optimizes over the worst-possible groups,
    \item distributionally robust MDP (\textbf{DR}), which holds a belief over MDPs without the hierarchical task structure and optimizes over the worst-possible belief distribution,
    \item robust MDP (\textbf{R}), which is a robust version of distributionally robust MDP and optimizes over the worst-possible MDP.
\end{itemize}

\vspace{-0.1in}
\paragraph{Optimal policy for R.} \textbf{R} desires robustness over the worst possible MDPs. We can see that the worst possible MDP is $m_1$ since the reward when choosing $a_0$ or $a_1$ in $m_1$ is consistently smaller than the rewards when in $m_0$. Since the optimal policy for $m_1$ is selecting $a_1$, the optimal policy for \textbf{R} is $a_1$. 

\vspace{-0.1in}
\paragraph{Optimal policy for GR.} \textbf{GR} desires robustness over the worst-possible mixtures. The value for selecting $a_0$ under mixture $z_0$ is $V(a_0|z_0) = 22*0.8=17.6$. Similarly, $V(a_1|z_0) = 5$, $V(a_0|z_1) = 0$ and $V(a_1|z_1) = 5$. Assume the agent has a stochastic policy, $\pi(p) = [p, 1-p]$, The value of the policy under mixture $z_0$ is $V(\pi(p), z_0) = 0.8*(22p+5*(1-p))+ 0.2*(0*p +5*(1-p)) = 12.6p+5$. The value of the policy under mixture $z_1$ is $V(\pi(p), z_1) = 0.5*(5p+0.0*(1-p))+ 0.5*(5p +0*(1-p)) = 5p$. Since $V(\pi( p), z_1) < V(\pi(p), z_0), \forall  p\in [0,1] $. The worst possible mixture is thus $z_1$ and the optimal policy for \textbf{GR} is $a_1$.

\vspace{-0.1in}
\paragraph{Optimal policy for DR.} \textbf{DR} desires robustness over the worst possible belief distribution over MDPs. The nominal $m$-level belief distribution is $b(m) = [0.4, 0.6]$, which is mapped from current $z$-level belief $b(z) = [0.5, 0.5]$.
Considering that there always exists one deterministic policy $\pi$ as the optimal policy for each belief distribution $b'(m)$, we directly analyze the value of the two actions with perturbed belief $\hat{b}(m)$. When the deterministic policy puts all mass on action $a_1$, perturbing belief doesn't affect the resulting value estimates since each $m$ has the same reward 5 when selecting $a_1$. Therefore the value of $a_1$ is always 5.
When the deterministic policy puts all mass on action $a_0$, the worst possible belief decreases the weight of $m_0$ by $\xi$, which is the maximum attack the adversary can apply. In this worst case, the value estimates of $a_1$ is $\hat{V} = (0.4-\xi)*22=4.4 < 5$. Therefore the optimal policy is $a_1$.

Similar results can be derived with the value function.
Formally, given $\epsilon \in [-\xi, \xi] = [-0.2, 0.2]$, $\pi(a_0) = p, \pi(a_1) = 1-p$, we want to solve the following optimization problem
\begin{align*}
    \max_{p}\min_{\epsilon}V(\pi(p), \gC_{b(m), \xi}) &= \max_{p}\min_{\epsilon}(0.4-\epsilon)[22p + 5(1-p)] + (0.6+\epsilon)[0p+5(1-p)] \\
    & = \max_{p}\min_{\epsilon}-22p\epsilon + 3.8p+5
\end{align*}
Since $\frac{\partial }{\partial  \epsilon} V(\pi(p), \gC_{b(m), \xi}) =-22p, p\in [0,1]$, we have $\argmin_{\epsilon}V(\pi(p), \gC_{b(m), \xi}) = 0.2$. 
\begin{align*}
    \max_{p}\min_{\epsilon}V(\pi(p), \gC_{b(m), \xi}) 
    & = \max_{p}-0.6p+5
\end{align*}
Therefore when $p=0$, the value is maximized. It shows that the optimal policy is $a_1$.

\vspace{-0.1in}
\paragraph{Optimal policy for GDR.} 
\textbf{GDR} instead desires robustness over the worst possible belief distribution over contexts. Similar to the analysis for DR, the value estimate of $a_1$, $\hat{V}(a_1)$, is always equal to 5 regardless of the perturbed $\hat{b}(z)$. 
Now we need to investigate the value when selecting deterministic policy as $a_0$. The weight on $z_0$ in the perturbed belief lies in range $\hat{b}(z_0) \in [0.5-\xi, 0.5+\xi]=[0.3, 0.7]$. The value estimate for $a_0$ is thus $\hat{V}(a_0) = \hat{b}(z_0)*0.8*22=17.6\hat{b}(z_0) \in [5.28, 12.32]$. Since the lower bound is larger than the value of $\hat{V}(a_1)=5$, the optimal policy for GDR is $a_0$.

Similarly, we can also write out the value function and the optimization problem.
\begin{align*}
    &\max_{p}\min_{\epsilon}V(\pi(p), \gC_{b(z), \xi})\\
    = &\max_{p}\min_{\epsilon}(0.5-\epsilon)[0.8*(22p+5(1-p))+0.2*(0p+5(1-p))] + (0.5+\epsilon)[0p+5(1-p)] \\
    = & \max_{p}\min_{\epsilon}-17.6p\epsilon + 3.8p+5
\end{align*}
Since $\frac{\partial }{\partial  \epsilon} V(\pi(p), \gC_{b(z), \xi}) =-17.6p, p\in [0,1]$, we have $\argmin_{\epsilon}V(\pi(p), \gC_{b(z), \xi}) = 0.2$. 
\begin{align*}
    \max_{p}\min_{\epsilon}V(\pi(p), \gC_{b(z), \xi}) 
    & = \max_{p}0.28p+5
\end{align*}
Therefore when $p=1$, the value is maximized. It shows that the optimal policy is $a_0$.

To sum up, the Hierarchical Latent Bandit example shows that our proposed GDR-MDP has the potential to find a less conservative policy compared with other robust formulations.

\clearpage

\section{Proofs}
\label{sec:appendix_proof}
\subsection{Proofs for Section~\ref{sec:convergence}: Convergence of GDR-MDP in Infinite-horizon Case}
\label{sec:proof_convergence}
\vspace{-0.1in}
This section proves the convergence of GDR-MDP in the infinite-horizon case. We first prove the Bellman expectation equation and Bellman optimality equation in Section~\ref{sec:appendix_bellman_proof}. We then show the contraction operator build on the Bellman optimality equation is a contraction operator in Section~\ref{sec:appendix_proof_contraction}. Finally, we show the convergence of GDR-MDP in Section~\ref{sec:appendix_proof_convergence}.
\subsubsection{Proofs for Proposition~\ref{prop_bellman}}
\label{sec:appendix_bellman_proof}
\vspace{-0.1in}

We provide the proof for the Bellman expectation equation as follows.
Starting from the definition of $V_t^{\pi}(b_t, s_t)$, we first separate the elements at time step $t$ from future timesteps.
We then find that the elements related to future timesteps starting from step $t+1$ could be aggregated to the group distributionally robust value at step $t+1$. 
\begin{align*}
    V_t^{\pi}(b_t, s_t) = & \min_{\hat{b}_{t:T} \in \gC_{b_{t:T}}} \mathbb{E}_{\hat{b}_{t:T}(z)} \mathbb{E}_{\mu_{z}(m)} \mathbb{E}_{m}^{\pi_{t:T}} \big[  \sum_{n=t}^{T} \gamma^{n-t} r_n  \vert b_t, s_t \big] \\
    = & \min_{\hat{b}_{t:T} \in \gC_{b_{t:T}}} \mathbb{E}_{\hat{b}_{t:T}(z)} \mathbb{E}_{\mu_{z}(m)} \mathbb{E}_{m}^{\pi_{t:T}} \big[  \{r_t + \gamma \sum_{n=t+1}^{T} \gamma^{n-t-1} r_n \} \vert b_t, s_t \big] \\
    = & \min_{\hat{b}_{t} \in \gC_{b_{t}}} \mathbb{E}_{\hat{b}_{t}(z)} \mathbb{E}_{\mu_{z}(m)} \mathbb{E}_{m}^{\pi_{t}} \big[  \{r_t + \gamma \cdot \\
    &\min_{\hat{b}_{t+1:T} \in \gC_{b_{t+1:T}}} \mathbb{E}_{\hat{b}_{t+1:T}(z)} \mathbb{E}_{\mu_{z}(m)} \mathbb{E}_{m}^{\pi_{t+1:T}}  \big[ \sum_{n=t+1}^{T} \gamma^{n-t-1} r_n \big] \} \vert b_t, s_t \big] \\
    = & \min_{\hat{b}_{t} \in \gC_{b_{t}}} \mathbb{E}_{\hat{b}_{t}(z)} \mathbb{E}_{\mu_{z}(m)} \mathbb{E}^{\pi_{t}} \big[  \{ \mathbb{E}_{\ermR_m} [r_t] + \gamma \cdot \sum_{s_{t+1}} \ermP_m(s_{t+1}| s_t, a_t) \cdot \\
    &\min_{\hat{b}_{t+1:T} \in \gC_{b_{t+1:T}}} \mathbb{E}_{\hat{b}_{t+1:T}(z)} \mathbb{E}_{\mu_{z}(m)} \mathbb{E}_{m}^{\pi_{t+1:T}}  \big[ \sum_{n=t+1}^{T} \gamma^{n-t-1} r_n \big] \} \vert b_t, s_t \big] \\
    = & \min_{\hat{b}_{t} \in \gC_{b_{t}}} \mathbb{E}_{\hat{b}_{t}(z)} \mathbb{E}_{\mu_{z}(m)} \mathbb{E}^{\pi_{t}} \big[  \mathbb{E}_{\ermR_m} [r_t] + \gamma \cdot \sum_{s_{t+1}} \ermP_m(s_{t+1}| s_t, a_t) \cdot \\
    &\min_{\hat{b}_{t+1:T} \in \gC_{b_{t+1:T}}} \mathbb{E}_{\hat{b}_{t+1:T}(z)} \mathbb{E}_{\mu_{z}(m)} \mathbb{E}_{m}^{\pi_{t+1:T}}  \big[ \sum_{n=t+1}^{T} \gamma^{n-t-1} r_n \vert b_{t+1}=SE(b_t, s_t), s_{t+1}\big] \}  \big] \\
    = & \min_{\hat{b}_{t} \in \gC_{b_{t}}} \mathbb{E}_{\hat{b}_{t}(z)} \mathbb{E}_{\mu_z(m)} \mathbb{E}^{\pi_t} \Big[ \mathbb{E}_{\ermR_{m}}[r_t] +  \gamma \sum_{s_{t+1}} \ermP_m(s_{t+1}| s_t, a_t) V^{\pi}_{t+1}(b_{t+1}, s_{t+1})   \Big].
\end{align*}

Therefore, the Group Distributionally Robust Bellman expectation equation is
\begin{align*}
    V_t^{\pi}(b_t, s_t) &= \min_{\hat{b}_{t} \in \gC_{b_{t}}} \mathbb{E}_{\hat{b}_{t}(z)} \mathbb{E}_{\mu_z(m)} \mathbb{E}^{\pi_t} \Big[ \mathbb{E}_{\ermR_{m}}[r_t] +  \gamma \sum_{s_{t+1}} \ermP_m(s_{t+1}| s_t, a_t) V^{\pi}_{t+1}(b_{t+1}, s_{t+1})   \Big].   
\end{align*} 

\begin{proposition}
The Group Distributionally Robust Bellman optimality equation is
\begin{align*}
    V_t^{\pi^{\star}}(b_t, s_t) &= \max_{\pi_{t}} \min_{\hat{b}_{t} \in \gC_{b_{t}}} \mathbb{E}_{\hat{b}_{t}(z)} \mathbb{E}_{\mu_z(m)} \mathbb{E}^{\pi_t} \Big[ \mathbb{E}_{\ermR_{m}}[r_t] +  \gamma \sum_{s_{t+1}} \ermP_m(s_{t+1}| s_t, a_t) V^{\pi^{\star}}_{t+1}(b_{t+1}, s_{t+1})   \Big]. 
\end{align*} 
\end{proposition}

Following a similar process, we could also prove the Bellman optimality equation as follows.
\begin{align*}
    V_t^{\pi^{\star}}(b_t, s_t) = & \max_{\pi_{t:T}}\min_{\hat{b}_{t:T} \in \gC_{b_{t:T}}} \mathbb{E}_{\hat{b}_{t:T}(z)} \mathbb{E}_{\mu_{z}(m)} \mathbb{E}_{m}^{\pi_{t:T}} \big[  \sum_{n=t}^{T} \gamma^{n-t} r_n  \vert b_t, s_t \big] \\
    = & \max_{\pi_{t:T}}\min_{\hat{b}_{t:T} \in \gC_{b_{t:T}}} \mathbb{E}_{\hat{b}_{t:T}(z)} \mathbb{E}_{\mu_{z}(m)} \mathbb{E}_{m}^{\pi_{t:T}} \big[  \{r_t + \gamma \sum_{n=t+1}^{T} \gamma^{n-t-1} r_n \} \vert b_t, s_t \big] \\
    = & \max_{\pi_{t}}\min_{\hat{b}_{t} \in \gC_{b_{t}}} \mathbb{E}_{\hat{b}_{t}(z)} \mathbb{E}_{\mu_{z}(m)} \mathbb{E}_{m}^{\pi_{t}} \big[  \{r_t + \gamma \cdot \\
    &\max_{\pi_{t+1:T}}\min_{\hat{b}_{t+1:T} \in \gC_{b_{t+1:T}}} \mathbb{E}_{\hat{b}_{t+1:T}(z)} \mathbb{E}_{\mu_{z}(m)} \mathbb{E}_{m}^{\pi_{t+1:T}}  \big[ \sum_{n=t+1}^{T} \gamma^{n-t-1} r_n \big] \} \vert b_t, s_t \big] \\
    = & \max_{\pi_{t}} \min_{\hat{b}_{t} \in \gC_{b_{t}}} \mathbb{E}_{\hat{b}_{t}(z)} \mathbb{E}_{\mu_{z}(m)} \mathbb{E}^{\pi_{t}} \big[  \{ \mathbb{E}_{\ermR_m} [r_t] + \gamma \cdot \sum_{s_{t+1}} \ermP_m(s_{t+1}| s_t, a_t) \cdot \\
    & \max_{\pi_{t+1:T}}\min_{\hat{b}_{t+1:T} \in \gC_{b_{t+1:T}}} \mathbb{E}_{\hat{b}_{t+1:T}(z)} \mathbb{E}_{\mu_{z}(m)} \mathbb{E}_{m}^{\pi_{t+1:T}}  \big[ \sum_{n=t+1}^{T} \gamma^{n-t-1} r_n \big] \} \vert b_t, s_t \big] \\
    = & \max_{\pi_{t}}\min_{\hat{b}_{t} \in \gC_{b_{t}}} \mathbb{E}_{\hat{b}_{t}(z)} \mathbb{E}_{\mu_{z}(m)} \mathbb{E}^{\pi_{t}} \big[  \mathbb{E}_{\ermR_m} [r_t] + \gamma \cdot \sum_{s_{t+1}} \ermP_m(s_{t+1}| s_t, a_t) \cdot \\
    &\max_{\pi_{t+1:T}} \min_{\hat{b}_{t+1:T} \in \gC_{b_{t+1:T}}} \mathbb{E}_{\hat{b}_{t+1:T}(z)} \mathbb{E}_{\mu_{z}(m)} \mathbb{E}_{m}^{\pi_{t+1:T}}  \big[ \sum_{n=t+1}^{T} \gamma^{n-t-1} r_n \vert b_{t+1}=SE(b_t, s_t), s_{t+1}\big] \}  \big] \\
    = & \max_{\pi_{t}} \min_{\hat{b}_{t} \in \gC_{b_{t}}} \mathbb{E}_{\hat{b}_{t}(z)} \mathbb{E}_{\mu_z(m)} \mathbb{E}^{\pi_t} \Big[ \mathbb{E}_{\ermR_{m}}[r_t] +  \gamma \sum_{s_{t+1}} \ermP_m(s_{t+1}| s_t, a_t) V^{\pi^{\star}}_{t+1}(b_{t+1}, s_{t+1})   \Big].
\end{align*}

Therefore, the Group Distributionally Robust Bellman optimality equation is
\begin{align*}
    V_t^{\pi^{\star}}(b_t, s_t) &= \max_{\pi_{t}} \min_{\hat{b}_{t} \in \gC_{b_{t}}} \mathbb{E}_{\hat{b}_{t}(z)} \mathbb{E}_{\mu_z(m)} \mathbb{E}^{\pi_t} \Big[ \mathbb{E}_{\ermR_{m}}[r_t] +  \gamma \sum_{s_{t+1}} \ermP_m(s_{t+1}| s_t, a_t) V^{\pi^{\star}}_{t+1}(b_{t+1}, s_{t+1})   \Big].   
\end{align*}

\subsubsection{Proof for Lemma~\ref{lemma:contraction}}
\label{sec:appendix_proof_contraction}

Let $\mathcal{V}$ refer to a set of real-valued bounded functions on $\Delta^{Z-1} \times \mathcal{S}$ and $\mathcal{L}V(b, s): \mathcal{V} \rightarrow \mathcal{V}$ refer to the Bellman operator defined as 
\begin{align*}
    \mathcal{L}V(b, s) =\max_{\pi \in \Pi} \min_{\hat{b} \in \gC_{b}} \mathbb{E}_{\hat{b}(z)} \mathbb{E}_{\mu_z(m)} \mathbb{E}^{\pi} \Big[ \mathbb{E}_{\ermR_{m}}[r] +\gamma \sum_{s'} \ermP_m(s'| s, a) V^{\pi}(SE(b, s), s)   \Big]. 
\end{align*}

Now we start the proof to show that the Bellman operator above is a contraction operator. For notation simplicity, let
\begin{align*}
    \mathcal{L}_{\hat{b}}^{\pi}=
    \mathbb{E}_{\hat{b}(z)} \mathbb{E}_{\mu_z(m)} \mathbb{E}^{\pi} \Big[ \mathbb{E}_{\ermR_{m}}[r] +\gamma \sum_{s'} \ermP_m(s'| s, a) V^{\pi}(SE(b, s), s)   \Big], \text{ and } 
    \mathcal{L}V(b, s) =\max_{\pi \in \Pi} \min_{\hat{b} \in \gC_{b}} \mathcal{L}_{\hat{b}}^{\pi}.
    \label{eq:Bellman_operator_simplified}
\end{align*}

Given arbitrary $U, V \in \mathcal{B}$ and based on the definition of the operator $\mathcal{L}V$ above, $\mathcal{L}U, \mathcal{L}V$ are real-valued and bounded.

Let $(b_{U}, \pi_{U})$ and $(b_{V}, \pi_{V})$ be the saddle points for $\mathcal{L}U$ and $\mathcal{L}V$, respectively. 

Observe that, $\mathcal{L}_{b_{ U}}^{\pi_{ U}}U(b, s) \leq \mathcal{L}_{b_{ V}}^{\pi_{ U}}U(b, s)$ and $\mathcal{L}_{b_{ V}}^{\pi_{ V}}V(b, s) \geq \mathcal{L}_{b_{ V}}^{\pi_{ U}}V(b, s)$.

\begin{align*}
    & \ \| \mathcal{L}U(b, s) - \mathcal{L}V(b,s) \|_{\infty} \\
    =& \ \| \mathcal{L}_{b_{ U}}^{\pi_{ U}}U(b, s) - \mathcal{L}_{b_{ V}}^{\pi_{ V}}V(b, s) \|_{\infty} \\
    \leq & \ \|  \mathcal{L}_{b_{ V}}^{\pi_{ U}}U(b, s) - \mathcal{L}_{b_{ V}}^{\pi_{ U}}V(b, s) \|_{\infty} \\
    = & \ \| \mathcal{L}_{b_{ V}}^{\pi_{ U}} ( U(b, s) - V(b,s)) \|_{\infty} \\
    =& \ \|  \mathbb{E}_{\hat{b}(z)} \mathbb{E}_{\mu_z(m)} \mathbb{E}^{\pi}  \Big[  \gamma \sum_{s'} \ermP_m(s'| s, a) \cdot ( U(SE(b, s, \mu), s)  - V(SE(b, s, \mu), s)  ) \Big] \|_{\infty} \\
    \leq & \ \gamma \mathbb{E}_{\hat{b}(z)} \mathbb{E}_{\mu_z(m)} \mathbb{E}^{\pi}  \Big[   \sum_{s'} \ermP_m(s'| s, a) \cdot  \| U(SE(b, s, \mu), s)  - V(SE(b, s, \mu), s)  \|_{\infty} \Big] \\
    \leq & \ \gamma \| U(SE(b, s, \mu), s)  - V(SE(b, s, \mu), s)  \|_{\infty} \\
    = & \ \gamma \| U(b', s)  - V(b', s)  \|_{\infty}.
\end{align*}
Considering that $ 0< \gamma < 1$, we conclude that $\mathcal{L}V(b,s)$ is a contraction operator on complete metric space $(\mathcal{V}, \| \cdot \|_{\infty})$.

\subsubsection{Proof for Theorem~\ref{theorem_infinite_convergence}}
\label{sec:appendix_proof_convergence}
Since $\mathcal{L}V(b,s)$ is a contraction operator based on Lemma~\ref{lemma:contraction}, we directly follow the Banach's Fixed-Point Theorem \cite{heyman2004stochastic} to show that (a) there exist a unique solution for $\mathcal{L}V_{\infty}(b, s) = V_{\infty}(b, s)$, and (b) the value function initiating from any value converge uniformly by iterative applying the Bellman update built in finite horizon case.

\subsection{Proofs for Section~\ref{sec:analysis}: Robustness Guarantee for GDR-MDP}
\label{sec:appendix_robustness}
In this section, we prove the robustness guarantee of our proposed GDR-MDP. We compare the GDR-MDP's optimal value with three different robust formulations. 
We achieve the comparison by studying how maintaining beliefs over mixtures affects the worst-possible outcome of the inner minimization problem and the resulting RL policy.
We study the worst-possible value via the relationships between ambiguity sets projected to the space of beliefs over MDPs.

\subsubsection{Ambiguity Set Projection and Set Relationships}

Recall that we consider a discrepancy-based ambiguity set defined based on total variance distance in Definition~\ref{def:tv_set}. Formally, the ambiguity set is
\begin{align*}
    \gC_{\nu_{\gX}, d_{TV}, \xi}(X) = \{ \nu'(X) : \sup_{X \in \gX}|\nu'(X) - \nu_{\gX}(X)| \leq \xi\},
\end{align*}
where $x \in \gX$ is the support, $\nu_{\gX}$ is the nominal distribution over $\gX$, and $\xi$ is the ambiguity set's size.

Define a column stochastic matrix $\mA = ((a_{ij})) \in \mathbb{R}^{M \times Z}, i \in [M], j \in [Z]$, where $ a_{ij} = \mu_{z=j}(m=i)$ represents a conditional probability equal to the $i$-th element of $\mu_{z=j}$ defined in GDR-MDP. 

Based on the total probability theorem, the matrix $\mA$ maps distributions over $\gZ$ to distributions over $\gM$. Formally, $ \forall \rvp(z) \in [0,1]^Z, \sum_{\gZ}\rvp(z) = 1$, there exists $\rvp(m) = \mA \rvp(z), \rvp(m) \in [0,1]^M, \sum_{m}\rvp(\gM) = 1$. 

We now define the operator that maps an ambiguity set over distribution for mixtures to an ambiguity set over distributions for MDPs.
\begin{definition}[Ambiguity Set Projection] \label{def:T_map}
The operator $T_{\mA}$ projects an ambiguity set for distributions over $\gZ$ to an ambiguity set for distributions over $\gM$, and
\begin{align*}
    T_{\mA}(\gC_{b(z), d, \xi}(z)) = \{ \rvp'(m): \rvp'(m) = \mA \rvp(z), \forall \rvp(z) \in \gC_{b(z), d, \xi}(z) \}.
\end{align*}

$\gC_{b(m), d, \xi}(m)$ is the ambiguity set for admissible distributions over supports $\gM$, where $b(m)$ is the nominal distribution. $d$ is the distance metric. $\xi$ is the set size and also the adversary's perturbation budget around the nominal distribution.
Similarly, $\gC_{b(z), d, \xi}(z)$ is the ambiguity set for admissible distributions over supports $\mathcal{Z}$. 
\end{definition} 

With the set projection operator $T_{\mA}$, we can derive the relationships between the projected ambiguity set $T_{\mA}(\gC_{b(z), d, \xi}(z))$ and the $\xi$-ambiguity set $\gC_{b(m), d, \xi}(m)$ which directly represents the model misspecifications over different MDPs.
We state the results in Proposision~\ref{prop:set_relationship}.

\begin{proposition}[Ambiguity Set Regularization with the Hierarchical Latent Structure]
\label{prop:set_relationship}
Consider two adversaries with the same attack budget $\xi$. One adversary perturbs the $z$-level distribution by selecting the worst possible distribution within $\gC_{b(z), d, \xi}(z)$ and the other perturbs the $m$-level distribution by selecting the worst possible distribution within $\gC_{b(m), d, \xi}(m)$. 
Given the nominal distribution for $\gZ$ as $b(z)$, we have the following statements hold:
\begin{enumerate}
    \item $b(m) = \mA b(z)$.
    \item $T_{\mA}(\gC_{b(z), d, \xi}(z)) \subseteq \gC_{b(m), d, \xi}(m)$. The $m$-level ambiguity set projected from a $z$-level $\xi$-ambiguity set is a subset of the $m$-level $\xi$-ambiguity set when directly perturbing $m$-level distributions. It means the hierarchical structure imposes extra regularization/constraints on the adversary.
\end{enumerate}
\end{proposition}

The second statement in Proposition~\ref{prop:set_relationship} shows that the hierarchical structure imposes extra regularization/constraints on the adversary by shrinking the ambiguity set.
The actual regularization reflected on the perturbed value of $(b,s)$ is related to the rank of the matrix $\mA$ and the loss function of downstream tasks (e.g. the transition models in the group of RL). 
The hierarchical latent structure in GDR-MDP can be viewed as a mixture model with random variables as $m \in [M]$ such that $\gM_m \in \gM$, and latent variables as $z \in [Z]$. 
The results in Proposition~\ref{prop:set_relationship} are applicable for general mixture models.

We now provide the proof for Proposition~\ref{prop:set_relationship} as follows.

\begin{proof}[Proof for Proposition~\ref{prop:set_relationship}]
Item (1) directly follows the definition of operator $T_{\mA}$ in Definition~\ref{def:T_map}.

Define the ambiguity sets based on Definition~\ref{def:tv_set}, where the cost function is the cost total variance distance. 
\begin{align*}
    \gC_{b(m), d_{TV}, \xi}(m) &= \{ \rvp(m) : \sup_{M \in \mathcal{M}}|\rvp(m) - b(m)| \leq \xi\},\\
    \gC_{b(z), d_{TV}, \xi}(z) &= \{ \rvp(z) : \sup_{Z \in \mathcal{Z}}|\rvp(z) - b(z)| \leq \xi\}
\end{align*}

Consider an arbitrary $\rvp'(m) \in T_{\mA}(\gC_{b(z), d, \xi}(z))$, there exists a distribution $\rvp(z) \in \gC_{b(z), d, \xi}(z)$, such that $\rvp'(m) = A\rvp(z)$. Therefore, 
\begin{align*}
     \rvp'(m) - b(m) = \mA \rvp(z) -b(m) = \mA \rvp(z) -\mA b(z) = \mA (\rvp(z) - b(z))
\end{align*}
Let $g = \rvp'(m) - b(m)$. Denote the $i$-th element of $g$ as $g_i, i \in [n]$. Let $a_i \in {[0,1]}^{1\times m}$ denote the $i$-th row of $A$. 

Considering that elements in $a_i$ are non-negative and lie in interval $[0,1]$, we have
\begin{align*}
    g_i &= a_i^T(\rvp(z) - b(z)) \\
    & \leq a_i^T(\rvp(z) - b(z))_{+} & \text{($(\cdot)_{+}$ is an operator that replaces negative elements with 0)}\\
    &\leq  \sum (\rvp(z) - b(z))_{+} & \text{(each element in $a_i$ is in $[0,1]$)}\\
    & = d_{TV}(\rvp(z), b(z)) \\
    & \leq \xi, \ \forall i \in [n].
\end{align*}

Similarly, we can prove $-g_i \leq \xi, \ \forall i \in [n]$. 
\begin{align*}
    -g_i &= -a_i^T(\rvp(z) - b(z)) = a_i^T(b(z) - \rvp(z)) \leq a_i^T(b(z) - \rvp(z))_{+} \\
    &\leq  \sum (b(z) - \rvp(z))_{+} = d_{TV}(\rvp(z), b(z)) \\
    &\leq \xi, \ \forall i \in [n].
\end{align*}

Therefore, we have elements in $g$ bounded by $\xi$: $|g_i| \leq \xi, \ \forall i \in [n]$.
\begin{align*}
    &|g_i| \leq \xi, \ \forall i \in [n] \\
    \Rightarrow & |\mA\rvp(z) - b(m)| \leq \xi, \ \forall z \in [Z] & \text{(because of the definition of $g_i$)}\\
    \Rightarrow & \sup_{z \in \mathcal{Z}} |\mA\rvp(z) - b(m)| \leq \xi \\
    \Rightarrow & T_{\mA}(\gC_{\mu_{\mathcal{Z}}, d, \xi}(z)) \subseteq \gC_{\mu_{\mathcal{M}}, d, \xi}(m).
\end{align*}
\end{proof}

\paragraph{Remark.} $\mA$ is not a stochastic row matrix, which makes the proof different from the contraction mapping proof in tabular RL settings where the transition matrix is a stochastic row matrix.

\subsubsection{Proof for Theorem~\ref{theorem:value_analysis}}

With the ambiguity set relationships in Proposition~\ref{prop:set_relationship}, we are now ready to prove Theorem~\ref{theorem:value_analysis}.

Recall that for notation simplicity, let
$U_{m}(\pi) = \mathbb{E}_m^{ \pi} \big[\sum_{t=1}^{T} \gamma^t r_t \big]$.
Let $\gC_{b(m), d_{TV}, \xi}(m)$ and $\gC_{b(z), d_{TV}, \xi}(z)$ denote the ambiguity sets for beliefs over MDPs $m$ and mixtures $z$, respectively. $b(m)$ and $b(z)$ satisfy $b(m) = \sum_{\gZ}\mu_z(m)b(z)$ and are the nominal distributions.
For any history-dependent policy $\pi \in \Pi$, its value function under different robust formulations are:
\begin{align*}
    V_{GDR}(\pi) &=  \min_{\hat{b}(z) \in \gC_{b(z), d_{TV}, \xi}(z)} \mathbb{E}_{\hat{b}(z)} \mathbb{E}_{\mu_z(m)}  [ U_{m}(\pi) ], 
    &
    V_{GR}(\pi) &= \min_{z \in [Z]} \mathbb{E}_{\mu_z(m)}  [ U_{m}(\pi) ], \\
    V_{DR}(\pi) &= \min_{\hat{b}(m) \in \gC_{b(m), d_{TV}, \xi}(m)} \mathbb{E}_{\hat{b}(m)}  [ U_{m}(\pi) ],
    &
    V_{R}(\pi) &=  \min_{m \in [M]}[ U_{m}(\pi) ].
\end{align*}

\begin{proof}[Proof for Theorem~\ref{theorem:value_analysis}]

First prove item (1) which is $V_{GDR}(\pi) \geq V_{GR}(\pi) \geq V_{R}(\pi)$: 

Given an arbitrary policy $\pi \in \Pi$, we have 
\begin{align*}
    V_{GDR}(\pi) &= \min_{\hat{b}(z) \in \gC_{b(z), d_{TV}, \xi}(z)} \mathbb{E}_{\hat{b}(z)} \mathbb{E}_{\mu_z(m)}  [ U_{m}(\pi) ]\\
    & \geq \min_{\hat{b}(z) \in \Delta_{Z}} \mathbb{E}_{\hat{b}(z)} \mathbb{E}_{\mu_z(m)}  [ U_{m}(\pi) ] \\
    & =\min_{z\in |Z|} \mathbb{E}_{\mu_z(m)}  [ U_{m}(\pi) ]\\
    & = V_{GR}(\pi)
\end{align*}
It means that with a nontrivial ambiguity set $\gC_{b(z), d_{TV}, \xi}(z)$, the distributionally robust value is more optimistic than the group robust formulation.
\begin{align*}
    V_{GR}(\pi) &= \min_{z\in |Z|} \mathbb{E}_{\mu_z(m)}  [ U_{m}(\pi) ] \\
    & \geq \min_{z \in [Z]} \min_{m \sim \mu_z(m)}  [ U_{m}(\pi) ] \\
    & \geq \min_{m \in [M]}  [ U_{m}(\pi) ] \\
    & = V_{R}(\pi_1)
\end{align*}
Therefore, we have $V_{GDR}(\pi) \geq V_{GR}(\pi) \geq V_{R}(\pi)$.

\paragraph{Remark} The belief robust method with $V_{GR}$ is compatible with a non-adaptive robust problem, where the policy of the decision maker is a Markov policy that only depends on the current state. In contrast, the belief distributionally robust method with $V_{GDR}$ corresponds to an adaptive robust problem, where the decision maker utilizes a history-dependent policy. In other words, it considers both the current state and the information gathered along with interactions. A similar argument but in a non-robust version is presented as Proposition 1. in \cite{steimle2021multi}. 

Now prove the inequality relationship in item (2) which is $V_{GDR}(\pi) \geq  V_{DR}(\pi)$:

Based on the projection operator in Definition~\ref{def:T_map}, we change the minimization over belief distribution on mixtures to an equivalent expression that has minimization over belief distribution on MDPs instead.
\begin{align*}
    V_{GDR}(\pi)& =  \min_{\hat{b}(z) \in \gC_{b(z), d_{TV}, \xi}(z)} \mathbb{E}_{\hat{b}(z)} \mathbb{E}_{\mu_z(m)}  [ U_{m}(\pi) ]\\
    & =  \min_{\hat{b}(z) \in \gC_{b(z), d_{TV}, \xi}(z)}  \mathbb{E}_{m \sim \sum_{z}\hat{b}(z)\mu_z(m)}  [ U_{m}(\pi) ] \\
    & = \min_{\hat{b}(m) \in T_{\mA}(\gC_{b(z), d, \xi}(z))} \mathbb{E}_{\hat{b}(m)}  [ U_{m}(\pi) ] & \text{(based on Definition~\ref{def:T_map})}
\end{align*}

Then with Proposition~\ref{prop:set_relationship}, which shows the set relationships, we have,
\begin{align*}
    V_{GDR}(\pi)&  = \min_{\hat{b}(m) \in T_{\mA}(\gC_{b(z), d, \xi}(z))} \mathbb{E}_{\hat{b}(m)}  [ U_{m}(\pi) ] \\
    & \geq \min_{\hat{b}(m) \in \gC_{b(m), d_{TV}, \xi}(m)} \mathbb{E}_{\hat{b}(m)}  [ U_{m}(\pi) ] & \text{(because of $T_{\mA}(\gC_{b(z), d, \xi}(z)) \subseteq \gC_{b(m), d, \xi}(m)$)}\\
    &=  V_{DR}(\pi).
\end{align*}
It shows that, in general, distributionally robust over high-level latent variable $z$ is more optimistic than that over low-level latent variable $m$. The hierarchical mixture model structure help regularize the strength of the adversary and generate less conservative policies than the flat model structure.

Therefore, we have the following inequalities hold: $V_{GDR}(\pi) \geq V_{GR}(\pi) \geq V_{R}(\pi)$ and $V_{GDR}(\pi) \geq  V_{DR}(\pi)$.
\end{proof}

\subsubsection{Proof for Theorem~\ref{theorem:optimal_policies}}
Based on Theorem~\ref{theorem:value_analysis}, we can derive the relationships between the optimal values for different formulations.

\begin{proof}[Proof for Theorem~\ref{theorem:optimal_policies}]
First prove that $V_{GDR}(\pi_{GDR}^{\star}) \geq V_{DR}(\pi_{DR}^{\star}) $.

Since $\pi_{GDR}^{\star}$ is the optimal policy for GDR-MDP, we have 
\begin{align*}
    V_{GDR}(\pi_{GDR}^{\star}) \geq V_{GDR}(\pi_{DR}^{\star}).
\end{align*}

Since $V_{GDR}(\pi) \geq  V_{DR}(\pi), \forall \pi$, base on Theorem~\ref{theorem:value_analysis}, we have 
\begin{align*}
    V_{GDR}(\pi_{DR}^{\star}) \geq  V_{DR}(\pi_{DR}^{\star}).
\end{align*}
Therefore we have
\begin{align*}
    V_{GDR}(\pi_{GDR}^{\star}) \geq V_{DR}(\pi_{DR}^{\star}).
\end{align*}

Following similar procedures, we prove that $V_{GDR}(\pi_{GDR}^{\star}) \geq V_{GR}(\pi_{GR}^{\star}) \geq V_{R}(\pi_{R}^{\star}) $.
\begin{align*}
    V_{GDR}(\pi_{GDR}^{\star}) & \geq V_{GDR}(\pi_{GR}^{\star}) & \text{(since $\pi_{GDR}^{\star}$ is the optimal policy for GDR-MDP)}
    \\
    & \geq V_{GR}(\pi_{GR}^{\star}) 
    & \text{(since $V_{GDR}(\pi) \geq V_{GR}(\pi), \forall \pi$ in Theorem~\ref{theorem:value_analysis})}
    \\
    & \geq V_{GR}(\pi_{R}^{\star})
    & \text{(since $\pi_{GR}^{\star}$ is the optimal policy for group robust MDP)}
    \\
    &\geq V_{R}(\pi_{R}^{\star}). 
    & \text{(since $V_{GR}(\pi) \geq V_{R}(\pi), \forall \pi$ in Theorem~\ref{theorem:value_analysis})}
\end{align*}
Therefore, we have shown the following inequalities hold: $V_{GDR}(\pi_{GDR}^{\star}) \geq V_{GR}(\pi_{GR}^{\star}) \geq V_{R}(\pi_{R}^{\star}) $ and $V_{GDR}(\pi_{GDR}^{\star}) \geq V_{DR}(\pi_{DR}^{\star}) $.
\end{proof}

\newpage

\section{Environment Details}
\label{sec:appendix_environments}

\subsection{Google Research Football}

Google Research Football (GRF) is a physics-based 3D soccer simulator for reinforcement learning. 
This domain presents additional challenges due to its AI randomness, large state-action spaces, and sparse rewards. 
The RL agent will control one active player on the attacking team at each step and can pass to switch control. The non-active players will be controlled by built-in AI. In our designed 3 vs. 2 tasks, there are three attacking players of a certain type and two defending players, including one player of a chosen type and a goalkeeper.

\begin{figure}[h]
    \centering
    \includegraphics[width=1.0\linewidth]{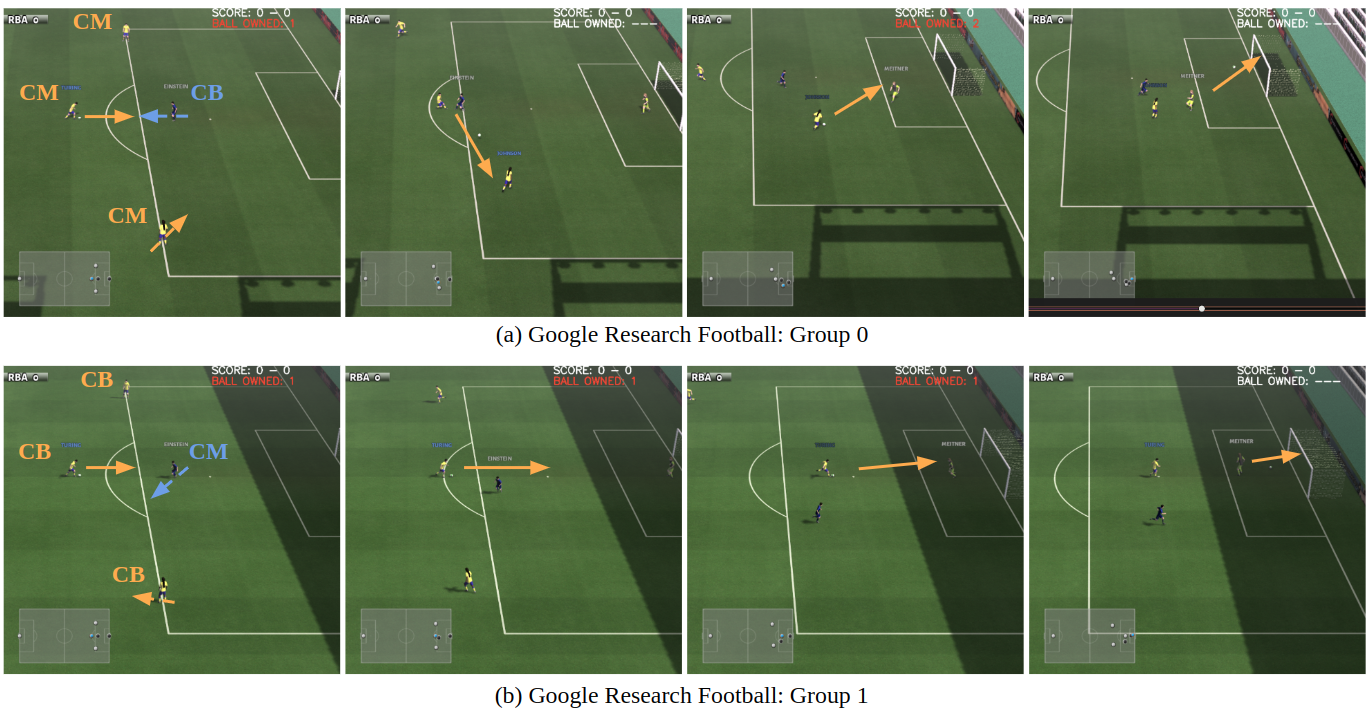}
    \vspace{-0.3in}
    \caption{This figure displays good solutions for tasks from two groups in GRF 3 vs. 2. The yellow solid arrow depicts the movement direction of the ball, the yellow dashed arrow depicts the movement direction of the attacking player on the wing, and the blue dashed arrow shows the movement direction of the defending player.}
    \label{fig:grf_vis}
    \vspace{0.2in}
\end{figure}

The dynamics of the 3 vs. 2 tasks are determined by the player types, including central midfield (CM) and centre back (CB), and player capability levels. The mixture index set $\gZ$ has a cardinality of two, $z=0$ and $z=1$, corresponding to CM vs. CB (with the goalkeeper) and  CB vs. CM (with the goalkeeper), respectively.
The built-in CM player tends to go into the penalty area when attacking and guard the player on the wing (physically left or right) when defending, while the CB player tends to guard the player in the middle when defending, and not directly go into the penalty area when attacking. 
Thus, different patterns of policies are required to solve the tasks from different groups. 
As shown in Figure \ref{fig:grf_vis}, in a CM-attacking-CB-defending task, a good solution is to first pass the ball to the player on the wing and then shoot. In a CB-attacking-CM-defending task, a good policy is to directly run into the penalty area and shoot.
To further encourage task diversity, we add some noisy actions to a run-into-penalty policy in a CM-attacking-CB-defending task, and to a pass-and-shoot policy in a CB-attacking-CM-defending task, when the controlled player faces high-intensity defense.

For the player capability level, we have two types of settings, players with 1.0 capability attacking while players with 0.7 capability defending (1.0 vs. 0.7), and players with 0.9 capability attacking while players with 0.6 capability defending (0.9 vs. 0.6). The strongest player has a capability level of 1.0. It is worth noting that these settings are more challenging than the original 3 vs. 2 task in GRF (1.0 vs. 0.6) in terms of capability level.
Detailed descriptions of the state and action space are shown in Table \ref{tab:obs_act_grf}.
\begin{table}[h]
\vspace{0.1in}
    \caption{Observation and action space in Google Research Football}
    \label{tab:obs_act_grf}
    \vspace{-0.1in}
    \centering
    \begin{tabular}{lll}
      \toprule
      Dim.  & Continuous Observation Space & range \\
      \midrule
      0-7   & $x, y$ positions of the attacking players (including the goalkeeper) & $[-1, 1]$      \\
      8-11  & $x, y$ positions of the defending players & $[-1, 1]$  \\
      12-19 & movements of the attacking players along $x, y$ directions & $[-1, 1]$  \\
      20-23 & movements of the defending players along $x, y$ directions & $[-1, 1]$  \\
      24-26 & $x, y, z$ positions of the ball & $[-\inf, \inf]$  \\
      27-29 & movements of the ball along $x, y, z$ directions & $[-1, 1]$  \\
      30-32 & $x, y, z$ rotation angles of the ball in radians & $[-\pi, \pi]$  \\
      33-35 & the one-hot encoding denoting the team controlling the ball & $\{0,1\}$  \\
      36-40 & the one-hot encoding denoting the player controlling the ball & $\{0,1\}$  \\
      41-42 & scores for each team (an episode terminates when any team scores) & $\{0,1\}$  \\
      43-46 & the one-hot encoding denoting the active player controlled by RL & $\{0,1\}$  \\
      47-56 & 10-elements vectors of 0s or 1s denoting whether a sticky action is active & $\{0,1\}$  \\
      \midrule
      Index     & Discrete Action Space    &  \\
      \midrule
      0     & idle  &   \\
      1     & run to the left, sticky action  &   \\
      2     & run to the top-left, sticky action &   \\
      3     & run to the top, sticky action  &   \\
      4     & run to the top-right, sticky action &   \\
      5     & run to the right, sticky action  &   \\
      6     & run to the bottom-right, sticky action  &   \\
      7     & run to the bottom, sticky action &   \\
      8     & run to the bottom-left, sticky action  &   \\
      9     & perform a long pass &   \\
      10    & perform a high pass  &   \\
      11    & perform a short pass  &   \\
      12    & perform a shot &   \\
      13    & start sprinting, sticky action  &   \\
      14    & reset current movement direction &   \\
      15    & stop sprinting  &   \\
      16    & perform a slide &   \\
      17    & start dribbling, sticky action  &   \\
      18    & stop dribbling &   \\
      \bottomrule
    \end{tabular}
\end{table}
\begin{table}[h]
\label{tab:appendix_GRF_task}
\caption{Detailed task descriptions for Google Research Football}
\vspace{-0.1in}
\centering
\small{
    \begin{tabular}{ccccc}
      \toprule
     \multirow{2}{*}{Task Index}  & Parameter 1 & Parameter 2 & \multirow{2}{*}{Group Index} & \multirow{2}{*}{Probability} \\ 
     & Player Type & Player Capability Level & &  \\ \midrule
    0 & CM vs. CB & 0.9 vs. 0.6 & 0 & 0.5 \\
    1 & CM vs. CB & 1.0 vs. 0.7 & 0 & 0.5 \\ \midrule
    2 & CB vs. CM & 0.9 vs. 0.6 & 1 & 0.5 \\
    3 & CB vs. CM & 1.0 vs. 0.7 & 1 & 0.5 \\
     \bottomrule
    \end{tabular}
}
\end{table}

\clearpage

\subsection{LunarLander}
\begin{figure}[h]
    \centering
    \includegraphics[width=1.0\linewidth]{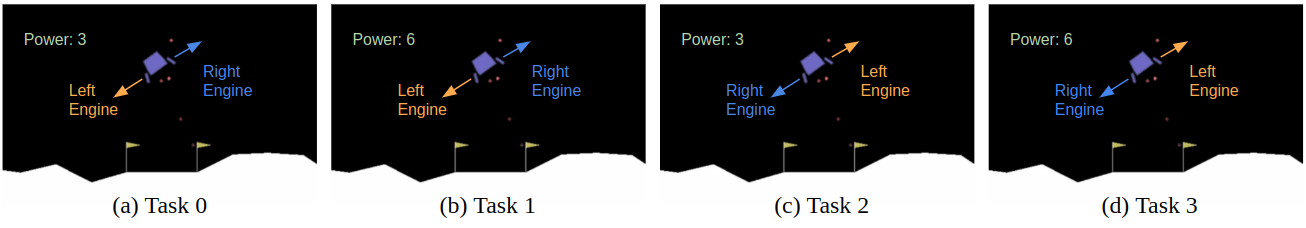}
    \vspace{-0.3in}
    \caption{LunarLander task visualization. Task 0 and task 1 belong to group 0 (normal mode). Tasks 2 and task 3 belong to group 1 (flip mode).}
    \label{fig:lunarlander_task_vis}
    \vspace{0.2in}
\end{figure}

We modify the LunarLander environment \cite{brockman2016openai} by changing the engine mode and engine power. The mixture index set $\gZ$ has a cardinality of two, $z=0$ and $z=1$, corresponding to two different engine operation modes, normal mode and left-right-flip mode, respectively. When in left-right-flip mode, the action turning on the left engine in normal mode will turn on the right engine instead, and the action turning on the right engine in normal mode will turn on the left instead. We visualize the tasks in Figure~\ref{fig:lunarlander_task_vis}.
The engine power has two choices which are 3.0 and 6.0. 
The MDP set $\gM$ has carnality four corresponding to four combinations of engine mode and engine power. 
Detailed descriptions of the state and action space are shown in Table~\ref{tab:obs_act_lunarlander}.

\begin{table}[h]
\label{tab:appendix_lunarlander_task}
\caption{Detailed task descriptions for LunarLander}
\vspace{-0.1in}
\centering
\small{
    \begin{tabular}{ccccc}
      \toprule
     \multirow{2}{*}{Task Index}  & Parameter 1 & Parameter 2 & \multirow{2}{*}{Group Index} & \multirow{2}{*}{Probability} \\ 
     & Engine Mode & Engine Power & &  \\ \midrule
     0 & Normal & 3.0 & 0 & 0.5 \\ 
     1 & Normal & 6.0 & 0 & 0.5 \\ \midrule
     2 & Flipped & 3.0 & 1 & 0.5 \\ 
     3 & Flipped & 6.0 & 1 & 0.5 \\
     \bottomrule
    \end{tabular}
}
\end{table}

\begin{table}[h]
    \vspace{0.1in}
    \caption{Observation and action space in LunarLander}
    \vspace{-0.1in}
    \label{tab:obs_act_lunarlander}
    \centering
    \small{
    \begin{tabular}{lll}
      \toprule
      Dim.  & Continuous Observation Space & range \\
        \midrule
        0     & $x$ position  & $[-\inf, \inf]$    \\
        1   & $y$ position & $[-\inf, \inf]$      \\
        2     & $x$ velocity  & $[-\inf, \inf]$     \\
        3   & $y$ velocity (relative): $x, y, v_x,v_y$ & $[-\inf, \inf]$  \\
        4    & angle  & $[-\pi, \pi]$     \\
        5 & angular velocity & $[-\inf, \inf]$   \\
        6 & if left leg contact with ground & $\{0,1\}$   \\
        7 & if right leg contact with ground & $\{0,1\}$   \\
        \midrule
        Index     & Discrete Action Space    &  \\
        \midrule
        0     & idle  &   \\
        1     & turn on left engine (normal mode)/Turn on right engine (left-right-flip mode)& \\
        2     & turn on main engine &  \\
        3     & turn on right engine (normal mode)/Turn on left engine (left-right-flip mode)&  \\
        \bottomrule
    \end{tabular}
    }
    \vspace{0.1in}
\end{table}

\clearpage

\subsection{HalfCheetah}

\begin{figure}[h]
    \centering
    \includegraphics[width=.5\linewidth]{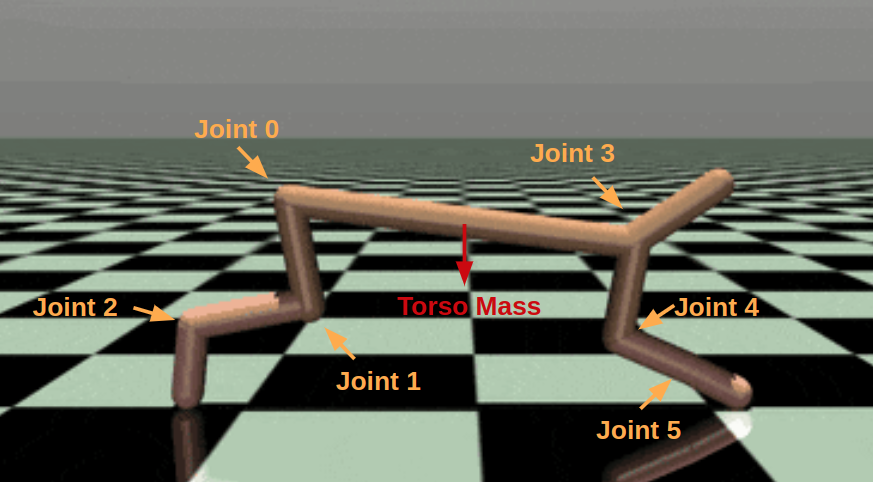}
    \vspace{-0.1in}
    \caption{HalfCheetah visualization.}
    \label{fig:halfcheetah_task_vis}
\end{figure}

We modify the joint failure and torso mass of HalfCheetah and build 18 tasks with different dynamics.
The joint failure has six choices which correspond to the 6 joints of HalfCheetah. For instance, when the joint failure index is 0, we cannot apply control torque (action) to joint 0.
The torso mass has three choices, which are 0.9, 1.0, and 1.1 times the original torso mass. 
We visualize the joint indexes in Figure~\ref{fig:halfcheetah_task_vis}.
Detailed descriptions of the state and action space are shown in Table~\ref{tab:obs_act_halfcheetah}.
\begin{table}[h]
    \caption{Observation and action space in HalfCheetah}
    \vspace{-0.1in}
    \label{tab:obs_act_halfcheetah}
    \centering
    \small{
    \begin{tabular}{ll}
      \toprule
      Dim.  & Continuous Observation Space \\
      \midrule
      0-8   & positional information    \\
      9-16   & velocity information  \\
      \midrule
      Dim     & Continuous Action Space    \\
      \midrule
      0-5     & control torque    \\
      \bottomrule
    \end{tabular}
    }
\end{table}
\begin{table}[h]
\label{tab:appendix_halfcheetah_task}
\caption{Detailed task descriptions for HalfCheetah}
\vspace{-0.1in}
\centering
\small{
\begingroup
    \begin{tabular}{ccccc}
      \toprule
     \multirow{2}{*}{Task Index}  & Parameter 1 & Parameter 2 & \multirow{2}{*}{Group Index} & \multirow{2}{*}{Probability} \\ 
      & Failure Joint & Torso Mass & & \\
     \midrule 
     0 & 0 & 0.9 & 0 & $1/3$ \\
     1 & 0 & 1.0 & 0 & $1/3$ \\
     2 & 0 & 1.1 & 0 & $1/3$ \\ \midrule
     3 & 1 & 0.9 & 1 & $1/3$ \\
     4 & 1 & 1.0 & 1 & $1/3$ \\
     5 & 1 & 1.1 & 1 & $1/3$ \\ \midrule
     6 & 2 & 0.9 & 2 & $1/3$ \\
     7 & 2 & 1.0 & 2 & $1/3$ \\
     8 & 2 & 1.1 & 2 & $1/3$ \\ \midrule
     9 & 3 & 0.9 & 3 & $1/3$ \\
     10 & 3 & 1.0 & 3 & $1/3$ \\
     11 & 3 & 1.1 & 3 & $1/3$ \\ \midrule
     12 & 4 & 0.9 & 4 & $1/3$ \\
     13 & 4 & 1.0 & 4 & $1/3$ \\
     14 & 4 & 1.1 & 4 & $1/3$ \\ \midrule
     15 & 5 & 0.9 & 5 & $1/3$ \\
     16 & 5 & 1.0 & 5 & $1/3$ \\
     17 & 5 & 1.1 & 5 & $1/3$ \\ 
     \bottomrule
     
    \end{tabular}
    \endgroup
}
\end{table}

\clearpage

\section{Implementation Details}
\label{sec:appendix_implementation_detail}

\vspace{-0.1in}
\paragraph{Trajectory rollout.} 
In both training and testing, we initialize the environment by sampling first a mixture and then an MDP realization. 
The sampled mixture and MDP are fixed throughout one episode.
In our environments with discrete mixtures and MDPs, we can represent the ground truth mixture index $\hat{z}$ with a one-hot vector $e_{\hat{z}}$, which is used in the pretraining phase of all baselines and in the whole training phase of baseline \textbf{G-Exact}. 
For baselines with belief module including \textbf{GDR}, \textbf{G-Belief}, \textbf{DR}, \textbf{State-R}, the actual mixture $\hat{z}$ and MDP weights $\mu(m|\hat{z})$ are unknown to the RL agent. 
Instead, the RL agent is given the number of possible mixtures $Z$ and is able to infer a belief over mixtures $b(z)$ based on a belief update function $SE$. A detailed algorithm for trajectory rollout is Algorithm~\ref{alg:rollout}.
For baseline \textbf{No-Belief}, we mask out the beliefs in the input by replacing them with zeros.

\paragraph{Belief update mechanism.} 
In our implementation (Section~\ref{sec:experiments}), we use the Bayesian update rule to update beliefs based on the interaction at each timestep.
At the beginning of each episode, we initialize a uniform belief distribution $b_0(i)=1/(|\gZ|), \forall i \in [|\gZ|]$. At timestep $t$, we update the belief as follows
\begin{align*}
    b_{t+1}(j) = \frac{b_t(j)L(j)}{\sum_{i \in [|\gZ|]} b_t(i)L(i)}, \forall j \in [|\gZ|],
\end{align*}
where $L$ represents the likelihood.
Let $\hat{z}$ denote the true mixture index for the episode.
We let the likelihood $L$ vector be a soft version of the actual one-hot mixture encoding $e_{\hat{z}}$. 

More concretely, at each time step, we first sample a noisy index $j$ where $j = \hat{z}$ with probability $\epsilon_l$ and $j$ is uniformly sampled from $[Z]$ otherwise.
The likelihood $L$ is a vector with dimension $|\gZ|$, and $\forall i \in [|\gZ|]$, the $i$-th element $L(i)$ is
\begin{align*}
    L(i) =  
    \begin{cases}
    l, & \text{if } i=j\\
    (1-l)/(|\gZ|-1), & \text{if }i\not=j
    \end{cases}
\end{align*}

There are lots of literature on accurate belief updates \cite{sokota2021fine}. In this work, we utilize a simple but controllable belief update mechanism above, which is more suitable for robustness evaluations since we could explicitly vary the hyperparameters. We leave a more sophisticated design of belief update mechanism for future work.

\paragraph{Belief noise level}
During robustness evaluation in Section~\ref{sec:discussion}, we control the belief noise level $\epsilon_{\hat{z}}$ which affects the likelihood $L$. More concretely, we add another layer of randomness on the estimate of $\hat{z}$.
Define the noisy mixture index at test-time as $z_{text}$, we have
\begin{align*}
    z_{test} = 
    \begin{cases}
    \hat{z} & \text{with probability }\epsilon_{\hat{z}} \\
    \text{a random index uniforms samples from }[|\gZ|], & \text{otherwise}
    \end{cases}
\end{align*}
During the robust evaluation, the likelihood $L_{test}$ is calculated based on $z_{test}$. More concretely, at each time step, we first sample a noisy index $j$ where $j = z_{test}$ with probability $\epsilon_l$ and $j$ is uniformly sampled from $[Z]$ otherwise. 
The likelihood and belief updates are as follows:
\begin{align*}
    L_{test}(i) =  
    \begin{cases}
    l, & \text{if } i=j\\
    (1-l)/(|\gZ|-1), & \text{if }i\not=j
    \end{cases}, \text{ and }
    b_{t+1}(j) = \frac{b_t(j)L_{test}(j)}{\sum_{i \in [|\gZ|]} b_t(i)L_{test}(i)}, \forall j \in [|\gZ|].
\end{align*}

\paragraph{Distributionally robust training with belief distribution over MDPs (DR)}
\textbf{DR} has an agent that takes the belief distribution $b(m)$ and state $s$ as inputs. \textbf{DR} uses the same belief updating rule as in \textbf{GDR} to update $b(z)$ at each timestep and then project $b(z)$ to $b(m)$ with $\mu_z(m)$.

This is a variant of our proposed Group Distributionally Robust DQN, which has a perturbed target taking $m$-level belief distribution as part of its input. 
Note that in \textbf{DR}, we still update $z$-level belief $b(z)$ based on the same belief updating function $SE$ as in \textbf{GDR}. However, in \textbf{DR}, for data pair $d$, the ambiguity set $\gC_{b'(m), d_{TV}, \xi}$ is centered at $b'(m)=T_{\mA}(b'(z))$ which is mapped from $b'(z)$. We also modify the fast gradient sign attack over $b(m)$ accordingly. We first sample $i \in [M]$ and apply attacks as $p(m)_j = p(m)_j + \alpha_{b} \cdot \sign ( \nabla_{p(m),j} V(p(m), s')), \ \forall j \not=i$ and $p(m)_i = p(m)_i - \sum_{j \not=i} p(m)_j$.  We iteratively apply the gradient sign attack to find $b^{adv}(m) =  \argmin_{ \substack{p(m) \in \gC_{b'(m), d_{TV}, \xi}} } \sum_{a\in \gA} Q_{\theta} ( p(m), s', a)$.

\subsection{GDR-PPO}

We represent the pseudo algorithm of GDR-PPO in Algorithm~\ref{alg:GDR_PPO}. We collect rollouts with un-perturbed beliefs and use the online rollouts to update the value network. To enhance the robustness to belief ambiguity, we tend to down-weight the probability of trajectories that may lead to large performance drops under the worst-possible belief within the ambiguity set.
Hence we construct a pseudo-advantage $\hat{A}$ by subtracting the performance drop $R_{drop}$ from the actual accumulated return. The worst-possible belief is calculated by FGSM.

\begin{algorithm}[h]
\SetAlgoLined
\SetNoFillComment
\caption{Group Distributionally Robust Training for GDR-PPO
}
\label{alg:GDR_PPO}
    \KwIn{Value-net $V_{\theta}(b(z), s)$, ambiguity set $\gC_{\cdot, d_{TV}, \xi}$, training episodes $N$ }
    {\bfseries Initialize} data buffer $\gD$ \;
    \For{$n=0$ {\bfseries to} $N$}{
     Rollout several episode with Algorithm~\ref{alg:rollout} and append data pairs to $\gD$ \;
    \If{Update Actor-net parameters}{
     Sample batch data from $\gD$ \;
    \For{Each trajectory in the batch}{
        Get advantage for the data pair at timestep $t$ $\hat{A}(b_t, s_t) = \sum_{t'=t}^{T-1} r_t - \Big(V(b_t, s_t) - \min_{ \substack{p(z) \in   \gC_{b_t(z), d_{TV}, \xi}}  } V_{\theta} ( p(z) , s_t) \Big) - V_{\theta}(b_t, s_t)$ \;
    }
    Update Actor-net with PPO \;
    }
    }    Return: Actor-net
\end{algorithm}

\subsection{Hyperparameters}
We show the hyperparameters for training Google Research Football, Lunarlander and Halfcheetah in Table~\ref{tab:param_grf}, Table~\ref{tab:param_lunarlander} and Table~\ref{tab:param_halfcheetah}, respectively. We select hyperparameters via grid search.
\begin{table}[h]
    \vspace{0.2in}
    \caption{Hyperparameters for the Google Research Football}
    \vspace{-0.1in}
    \label{tab:param_grf}
    \centering
    \small{
    \begin{tabular}{ll}
      \toprule
      reward decay & 0.997 \\
      net hidden structure & $[256, 256]$      \\
      net activation function & Tanh   \\
      learning rate & 0.00012 \\
      GAE ($\lambda$) & 0.95 \\
      clipping range & 0.115 \\
      entropy coefficient & 0.00155 \\
      value function coefficient & 0.5 \\
      number of environment steps per update & 8192 \\
      epoch & 10 \\
      adv budget & 0.2\\
      adv step size & 0.1 \\
      adv max step & 10 \\
      batch size & 256 \\
      \bottomrule
    \end{tabular}
    }
    \vspace{-0.1in}
\end{table}
\begin{table}[h]
    \vspace{0.2in}
    \caption{Hyperparameters for the LunarLander task}
    \vspace{-0.1in}
    \label{tab:param_lunarlander}
    \centering
    \small{
    \begin{tabular}{ll}
      \toprule
      reward decay & 0.95 \\
      net hidden structure & $[128, 128]$      \\
      net activation function & ReLU   \\
      value function learning rate & 0.01 \\
      value function learning rate decay & 0.999 \\
      epoch & 20 \\
      gradient steps per epoch & 5000 \\
      adv budget & 0.4\\
      adv step size & 0.02 \\
      adv max step & 50 \\
      batch size & 256 \\
      \bottomrule
    \end{tabular}
    }
    \vspace{-0.1in}
\end{table}
\begin{table}[h]
    \caption{Hyperparameters for the Halfcheetah}
    \vspace{-0.1in}
    \label{tab:param_halfcheetah}
    \centering
    \small{
    \begin{tabular}{ll}
      \toprule
      reward decay & 0.99 \\
      net hidden structure & $[256, 256]$      \\
      net activation function & ReLU   \\
      value function learning rate & 0.001 \\
      value function learning rate decay & 0.999 \\
      epoch & 200 \\
      gradient steps per epoch & 5000 \\
      adv budget & 0.2\\
      adv step size & 0.02 \\
      adv max step & 50 \\
      batch size & 256 \\
      \bottomrule
    \end{tabular}
    }
    \vspace{-0.1in}
\end{table}

\newpage

\section{Additional Ablation Study}
\label{sec:appendix_ablation}
\vspace{-0.1in}
In this section, we show how the ambiguity set size and the pretrain episodes affect the training stability and robustness of \textbf{DR}, which maintains a belief over MDPs. Compared with our proposed \textbf{GDR}, \textbf{DR} omits the hierarchical structure. 

\vspace{-0.1in}
\subsection{The effect of Ambiguity Set Size}
\begin{figure}[h]
    \centering
    \includegraphics[width=0.72\linewidth]{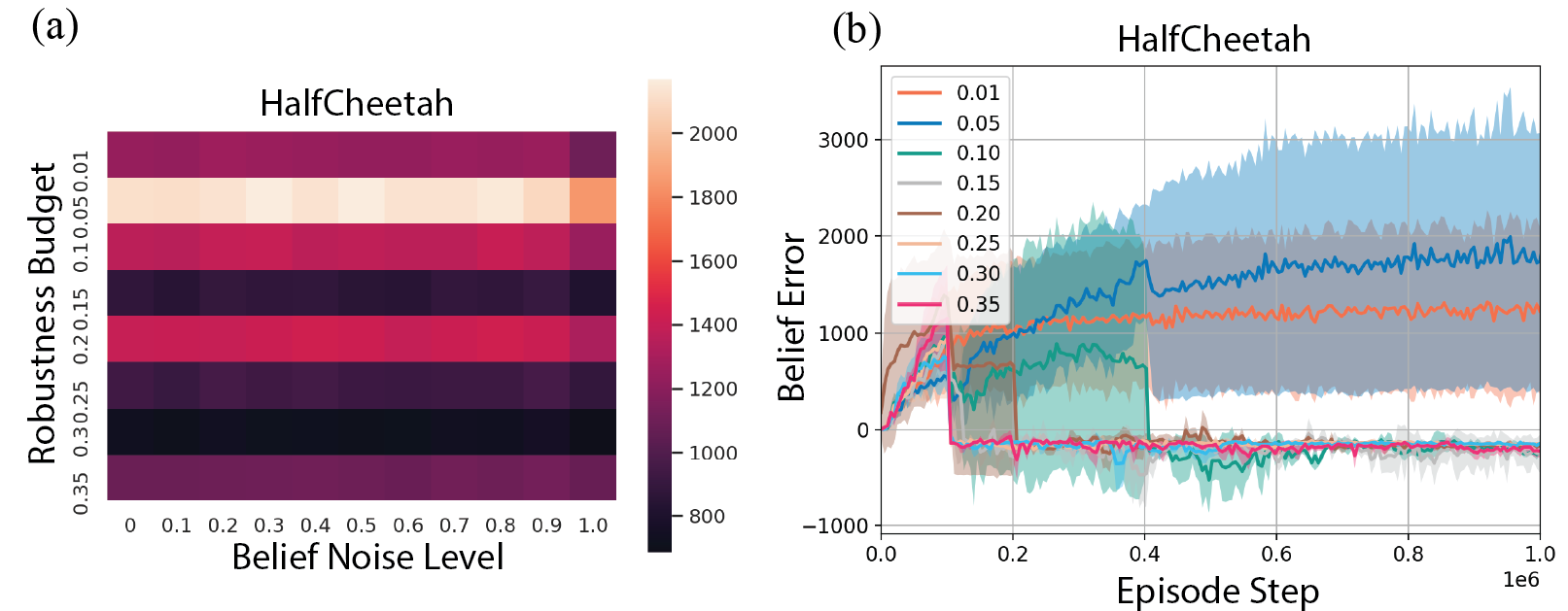}
    \vspace{-0.1in}
    \caption{Ablation study about the effect of ambiguity set budget on \textbf{DR}'s robustness and training stability. We choose the ambiguity set size among 0.01, 0.05, 0.1, 0.15, 0.2, 0.25, 0.3, and 0.35.}
    \label{fig:m_DR_budget_ablation}
\end{figure}

\vspace{-0.1in}
Figure~\ref{fig:m_DR_budget_ablation} shows the effect of the ambiguity set in HalfCheetah. All curves in Figure~\ref{fig:m_DR_budget_ablation} are pre-trained in the first 100000 episodes. 
With ambiguity set size 0.01 and 0.05, the \textbf{DR} does not crash and converge to a non-negative value. Comparing Figure~\ref{fig:m_DR_budget_ablation} (b) for \textbf{DR} with Figure~\ref{fig:training} (c) for \textbf{GDR}, we can conclude that \text{GDR} is less sensitive to the ambiguity set size along training since it converge to a non-negative value with a larger range of ambiguity set size. 
Comparing Figure~\ref{fig:ablation} (a) with Figure~\ref{fig:robustness} (c) for our proposed \textbf{GDR}, we can conclude that the hierarchical structure enhances the robustness to belief noise since the robustness performance for \textbf{GDR} consistently outperforms that of \textbf{DR} for different ambiguity set sizes.

\vspace{-0.1in}
\subsection{The effect of Pretrain Episodes}
\begin{figure}[h]
    \centering
    \includegraphics[width=0.72\linewidth]{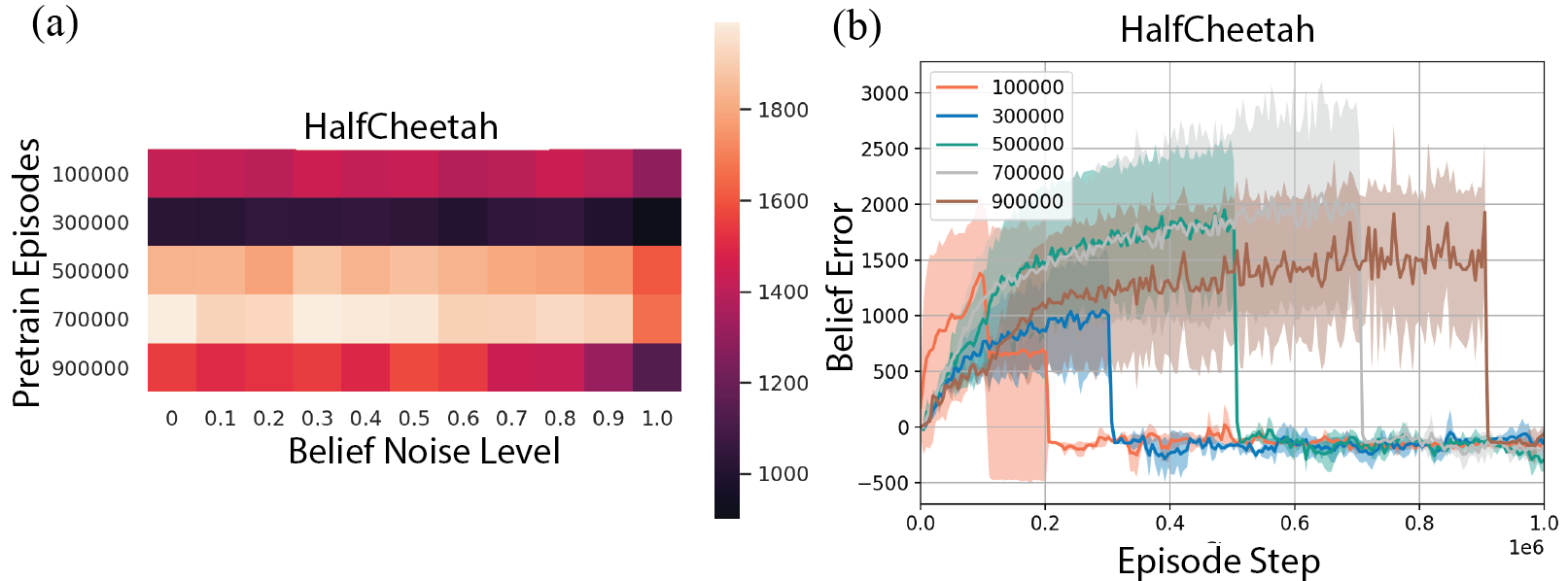}
    \vspace{-0.1in}
    \caption{Ablation study about the effect of pretrain episodes on \textbf{DR}'s robustness and training stability. We choose the number of pretrain episodes among 100000, 300000, 500000, 700000, and 900000.}
    \label{fig:m_DR_pretrain_ablation}
\end{figure}

\vspace{-0.1in}
Figure~\ref{fig:m_DR_pretrain_ablation} shows the effect of the pretrain episodes in HalfCheetah. All curves in Figure~\ref{fig:m_DR_pretrain_ablation} has an ambiguity set size 0.2. Figure~\ref{fig:m_DR_pretrain_ablation} shows that even pretraining for 900000 episodes, \textbf{DR} still will crash after the pretraining phase. It shows that \textbf{DR} is less sensitive to the pretraining episodes compared with the ambiguity set size. 
